\documentclass[msc,deptreport, ai]{infthesis} % Do not change except to add your degree (see above).
\usepackage{graphicx}
\usepackage[table]{xcolor}
\usepackage{amsmath}

\usepackage[utf8]{inputenc}
\usepackage[english]{babel}
\usepackage{hyperref}
\hypersetup{
    colorlinks=true,
    linkcolor=blue,
    filecolor=blue,      
    urlcolor=blue,
}
 
\graphicspath{ {./images/} }

\begin{document}
\begin{preliminary}

\title{Applying recent advances in Visual Question Answering to Record Linkage}

\author{Marko Smilevski}

\abstract{
Multi-modal Record Linkage is the process of matching multi-modal records from multiple sources that represent the same entity. This field has not been explored in research and we propose two solutions based on Deep Learning architectures that are inspired by recent work in Visual Question Answering. The neural networks we propose use two different fusion modules, the Recurrent Neural Network + Convolutional Neural Network fusion module and the Stacked Attention Network fusion module, that jointly combine the visual and the textual data of the records. The output of these fusion models is the input of a Siamese Neural Network that computes the similarity of the records. Using data from the Avito Duplicate Advertisements Detection dataset, we train these solutions and from the experiments, we concluded that the Recurrent Neural Network + Convolutional Neural Network fusion module outperforms a simple model that uses hand-crafted features. We also find that the Recurrent Neural Network + Convolutional Neural Network fusion module classifies dissimilar advertisements as similar more frequently if their average description is bigger than 40 words. We conclude that the reason for this is that the longer advertisements have a different distribution then the shorter advertisements who are more prevalent in the dataset. In the end, we also conclude that further research needs to be done with the Stacked Attention Network, to further explore the effects of the visual data on the performance of the fusion modules. 

%   This skeleton demonstrates how to use the \texttt{infthesis} style
%   for MSc dissertations in Artificial Intelligence, Cognitive Science,
%   Computer Science, Data Science, and Informatics. It also emphasises
%   the page limit, and that you must not deviate from the required
%   style.  The file \texttt{skeleton.tex} generates this document and
%   can be used as a starting point for your thesis. The abstract should
%   summarise your report and fit in the space on the first page.
}

\maketitle

\section*{Acknowledgements}
I would like to thank my supervisors Yoni Lev, Grant Galloway and Iain Murray for their advice, mentorship, patience and thorough feedback for my dissertation. I would also like to thank The University of Edinburgh and Amazon for providing an opportunity for me to work with mentors from the industry.

\tableofcontents
\end{preliminary}

\chapter{Introduction}
Record linkage is the process of joining records from multiple data sources that represents the same entity. Record linkage is used in domains where storing data is essential, therefore government agencies, the health sector, security agencies, online stores and other various organization would benefit immensely by any advances in this field. The task is commonly used for improving data quality, improving the order of items within databases and to reduce costs and give computational efficiency to data acquisition and data re-usage \cite{winkler2004methods}. 

All the research in record linkage has been done on textual data. Researchers have used various metrics to compute the similarity between textual fields and have proposed different machine learning methods to classify the output of these metrics \cite{Elmagarmid}. The problem with this approach is that for different domains, researches needed to use different similarity metrics and needed to design a lot of hand-crafted features, so the metric would be able to represent the similarity more accurately. This led to inefficient system development and systems that are not robust. In recent years Deep Learning has emerged as a solution to hand-crafting features, because of the ability of deep neural networks to extract features and knowledge from raw data \cite{Goodfellow-et-al-2016}. For record linkage, \cite{Mudgal:2018:DLE:3183713.3196926} proposes three architectures that are based on Recurrent Neural Networks and they outperform the state-of-the-art record linkage system (that uses classical machine learning methods) on raw text.

One thing that hasn't been researched in record linkage is the linkage of multi-modal records. The rise of the Internet has introduced more visual content which provides more information about the entities. Our motivation is to build on top of the success that \cite{Mudgal:2018:DLE:3183713.3196926} had with Deep Learning methods and we do this by proposing two Deep Learning architectures that will learn how to model multi-modal records and learn how to compute their similarity successfully. Another field that uses multi-modal data and combines them jointly is Visual Question Answering (VQA)\cite{antol2015vqa}. In VQA they use Deep Learning architectures to combine an image and a question and they use the output of the fusion module to predict the answer to the question. In this way, VQA relates to Record Linkage, because both problems need to jointly encode both the visual and textual data. 

The solutions that we propose consist of two parts: \textbf{two fusion modules} that combine the visual and textual information for each record in the pair separately and a \textbf{Siamese Neural Network} that computes the similarity between the outputs of the fusion modules. For the fusion modules we propose, two solutions that were inspired by the fusion modules introduced in \cite{antol2015vqa} and \cite{YangHGDS15}. The first proposed module is the \textbf{Recurrent Neural Network + Convolutional Neural Network fusion module} that uses recurrent neural networks to model the textual data and a pre-trained convolutional neural network that extracts features from images. The outputs of the networks are then fused by point-wise multiplication. An extension to this module is the \textbf{Stacked Attention Network fusion module}, that uses multiple attention layers to create a more refined encoding of the multi-modal data. To train the whole network we used the Avito Duplicate Ads Detection dataset \cite{avito}. This dataset hasn't been used in any previous research, so we also propose a simple baseline model that we will use to compare our two proposed solutions. The baseline model is a logistic regression model and for each pair of records, we compute two features: the Jaccard similarity coefficient between the textual data of each advertisement and the Euclidian distance between the feature vectors of the images associated with each advertisement.

We hypothesized that by increasing the complexity of the structure of the fusion module we would get better results. We were also interested in how the models perform when the textual data in the advertisements is long. Finally, we hypothesised that the Stacked Attention Network fusion module would be able to learn how to find similar regions in the images of similar advertisements.

For the \textbf{Recurrent Neural Network + Convolutional Neural Network fusion module} we did a small hyperparameter search and choose the model that performed best on the validation set. For the \textbf{Stacked Attention Network} we trained only one model because of time constraints.

From the experiments, we concluded that the \textbf{Recurrent Neural Network + Convolutional Neural Network} performs better than the baseline model. We failed to prove our hypothesis that the increase in fusion module complexity (upgrading the RNN + CNN module to the Stacked Attention Network module) would result in better model performance. We also failed to prove that the Stacked Attention Network solution can learn how to attend similar regions in the images of similar advertisements, due to time constraints. We did however found out that the hyperparameter setting used in \cite{antol2015vqa} and \cite{koch2015siamese} didn't work for the \textbf{Recurrent Neural Network + Convolutional Neural Network} solution. Another interesting discovery was that the \textbf{Recurrent Neural Network + Convolutional Neural Network} model, both struggle with advertisements that have very different description lengths and advertisements that have long descriptions. This is because the \textbf{Recurrent Neural Network + Convolutional Neural Network} model tends to overfit the distribution of pairs of advertisements that have an average description length smaller than 40 words. The distribution of these type of pairs consists of more similar pairs and it is different than the distribution of pairs of advertisements that have an average description length of more then 40 words.

The code used for this project is available at \url{https://github.com/msmilevski/nrl}.

The following paragraph outlines the structure for the rest of the document. In Chapter 2 we talk about the related work done in Record Linkage and provide the needed background for the proposed models. Section 2.1 explores different similarity metrics and machine learning methods (supervised and unsupervised) that are widely used in record linkage. Section 2.2 introduces the reader to Deep Learning, explains the connection between Deep Learning and Record Linkage and also gives a short introduction to Recurrent and Convolutional Neural Networks. Section 2.3 explores the link between Visual Question Answering and Record Linkage, explains what Visual Question Answering is in more detail and also introduces the concept of the Attention mechanism. Section 2.4 introduces Siamese Neural Networks. Chapter 3 is for the Methodology. In Section 3.1 we describe the dataset that we are using and all the preprocessing that we have done to it. Section 3.2 outlines the baseline model. Section 3.3 describes the Siamese Neural Network architecture in detail and provides an illustration of the architecture of the proposed framework. Section 3.4 describes the Recurrent Neural Network + Convolutional Neural Network fusion module and provides an illustration of its architecture. Section 3.5 explains the concept of the Stacked Attention Network, provides mathematical formulas to better understand the fusion module and a diagram of how the Stacked Attention Network looks when it uses one attention layer. Chapter 4 presents the results from the experiments. In Section 4.1 we introduce the evaluation metric we use to compare our models. Section 4.2 presents the results for the baseline model. Section 4.3 explains the training process for neural networks. Section 4.4 presents the results from the hyperparameter search for the Recurrent Neural Network + Convolutional Neural Network fusion module and we explain the thought process behind our experiments. Section 4.5 presents the results for the Stacked Attention Network fusion module. Section 4.6 compares all three models that we trained. It also looks into how well the models perform for different text lengths. Chapter 5 summarizes our work and introduces some possibilities for future work. 
\chapter{Background and related work}

\section{Record Linkage}
The goal of record linkage is to identify records in the same or different databases that refer to the same real-world entity, even if the records are not identical \cite{Elmagarmid}. The same task in other domains is referred to as entity matching, entity resolution or database deduplication. It is a crucial task for data integration and data cleaning and it has a massive financial impact on companies that store and work with large databases \cite{koudas2006record}. The general process of record linkage consists of four steps: data cleaning and standardisation, indexing, record pair comparison and similarity vector classification \cite{Elmagarmid}. The main task of data cleaning and standardisation is the conversion of raw input data into well defined, consistent forms, as well as the resolution of inconsistencies in the way information,  is represented and encoded \cite{5887335}. The indexing step aims to reduce the number of comparisons between the records in the databases by dismissing record pairs who are non-matching. This is done by techniques like blocking \cite{Elmagarmid}, which will not be covered in this thesis. 

The record pair comparison uses a variety of comparison functions (similarity metrics) appropriate to the content of the record fields. These similarity metrics are divided into four categories: Character-based, Token-based, Phonetic and Numeric similarity metrics \cite{Elmagarmid}. With character-based similarity metrics we measure the distance between fields on character level, with token-based we compare them on the word level and we use numeric similarity metrics if we compare numbers. Phonetic similarity metrics are used when two words are written in different languages, therefore phonetically they are similar but on character level, they would be very dissimilar. Widely used character-based similarity metrics are Edit-distance, Levenshtein distance and Jaro-Winkler distance, while for token-based similarity researchers have used metrics like TF-IDF cosine similarity, Jaccard coefficient and probabilistic analogues \cite{Elmagarmid}\cite{5887335}. The problem with these similarity metrics is that they are good for specific datasets and task, therefore making them not robust enough. For example, Edit-distance is suitable for common typing mistakes but it may be costly operation for large strings and problematic for specific domains \cite{koudas2006record}. Only a few attempts for comparative evaluations of some sub-approaches have been made in terms of evaluation of different string similarity metrics, but most published entity resolution evaluations focus on individual approaches and use diverse methodologies, measures, and test problems making it difficult to assess the quality of each approach, not to mention their comparative effectiveness and efficiency \cite{Kopcke}. Furthermore, \cite{Kopcke} find that matching product entities from online shops is not sufficiently solved with conventional approaches based on the similarity of attribute values because these approaches have poor classification performance. This means that matching records from online shops is a challenging resolution task.

The similarity metrics produce a numerical representation of the similarity between the entities. Non-learning systems have a pre-defined threshold \cite{Kopcke} which decides whether two records are similar. This can be problematic because the data may be noisy or the similarity metric might output values that are not linearly separable. Therefore the system might achieve better performance if it was able to learn the threshold. To create a system that can learn the threshold parameter the system should include machine learning methods in its structure. The similarity metrics produce a similarity coefficient/vector, that is the input of a classifier which decides whether the pair is matching or not. These classifiers are based on supervised and unsupervised machine learning methods. Supervised approaches aim at automating the process of record linkage. One of the more famous models is the Fellegi-Sunter model \cite{doi:10.1080/01621459.1969.10501049}, which approaches record linkage as a Bayesian inference problem. Its goal is to approximate the sets of similar and dissimilar item pairs by minimizing the number of matches that we are uncertain about. Other supervised solutions for record linkage include SVM \cite{Bilenko:2003:ANM:1137237.1137369} and the CART algorithm \cite{Cochinwala:2001:EDR:565996.565997} (decision trees) as the similarity vector classifier. However, supervised models require suitable labelled training data and providing such data requires manual effort. The drawback to these classifiers is that they are not robust enough, which means that when the data distribution changes slightly, new methods for feature extraction would be needed. As a solution to this problem, there are unsupervised methods for record linkage that propose different clustering techniques to group similar records together. The result of the clustering is that each cluster corresponds to a single distinct entity. One of the most cited solutions is Canopy clustering \cite{mccallum2000efficient} where the dataset is first divided into smaller subsets (canopies) by computing the similarity between pairs with an extremely inexpensive method. After the canopies are built, a standard clustering algorithm is applied to the canopies and they are clustered by using a more expensive distance (similarity) metric. Significant computation is saved by eliminating all of the distance comparisons among points that do not fall within a common canopy \cite{mccallum2000efficient}. Other clustering algorithms have been explored, like the Star clustering algorithm, Ricochet family of clustering algorithms, Cut Clustering, Articulation Point Clustering, Markov Clustering and Correlation Clustering and what \cite{hassanzadeh2009framework} concluded is that none of these clustering algorithms produces perfect clusters. Therefore \cite{hassanzadeh2009framework} propose that the best approach is to retrain your model more frequently and keep the important quantitative information produced by these algorithms. This only reinforces the idea that these solutions are not robust enough and their scalability would be problematic since frequent retraining is computationally expensive for large training sets. A more recent unsupervised method was proposed by \cite{Gruenheid} where they introduced the idea of incremental graph clustering technique for the record linkage task, meaning that the matching rules evolve through iterations.

The problem with unsupervised and supervised models is that there is no comparison between their performances and that they are mostly trained and evaluated on different datasets \cite{Elmagarmid}. For our research we used the Avito Duplicate Ads Detection dataset \cite{avito} and it contains labelled data about which pairs of advertisements (records) are similar and which are not. Because of this we decided to explore supervised modelling techniques. The current state-of-the-art solution for record linkage is Magellan which is an end-to-end EM (entity matching) system, that uses either supervised learning methods, rule-based methods or an ensemble of supervised learning and rule-based algorithms \cite{KondaDCDABLPZNP16}. All of the above-mentioned models work well with structured data, but classical machine learning methods may have difficulties matching textual instances because there are few meaningful features that we can create from raw text \cite{Mudgal:2018:DLE:3183713.3196926}. \cite{Kopcke} found that classical learning methods also perform poorly for the classification of matching product entities from online shops. Another issue is that our dataset contains images and a lot of hand-crafted feature engineering would be needed to capture the relevant visual features of each image.
\section{Deep Learning}
\subsection{Short introduction to Deep Learning}
Deep learning \cite{Goodfellow-et-al-2016} is an increasingly popular field in Artificial Intelligence. It is based on the idea of using a large number of perceptrons to approximate the distribution of the training data. Each perceptron consists of a weight that transforms the input (usually by multiplication) and an activation function that further transforms the weighted input into an output. By having weights and output, we can approximate the error that the perceptron makes and using backpropagation we can adjust the weights of the perceptron so that the output of the perceptron is a close to the real value. 
By using multiple perceptrons inline, we can create a network, commonly referred to as a fully connected neural network. If we stack two fully connected neural network, we create a Multi-Layer Perceptron. Multi-Layer Perceptrons are considered to be universal function approximators, hence the idea of Deep Learning is that by stacking more layers in-depth, the model would be able to approximate very complex functions. The input is usually referred to as the visible layer, and all the other layers in the deep architectures are called hidden layers. There are various types of layers \cite{Goodfellow-et-al-2016}, but the most notable, apart from the fully connected neural networks, are the recurrent neural networks and the convolutional neural networks, which will be discussed in the further sections. 
\subsection{Deep Learning and Record Linkage}
Deep learning models have had success in various Natural Language Processing tasks, like language modelling \cite{6424228} and machine translation \cite{bahdanau2014neural} amongst others. In their work, the main building block for modelling language (sentences) were recurrent neural networks. Because of this \cite{Mudgal:2018:DLE:3183713.3196926} proposed three different architectures based on recurrent neural networks for record linkage. These architectures outperform the Magellan system for entities that were consisted of raw and unfiltered textual fields, but they underperformed for structured entities. The data that we use to train our models is raw textual data with an additional image component. The results \cite{Mudgal:2018:DLE:3183713.3196926} encouraged us to use deep learning as the building block of our models. Therefore in the next subsections, we introduce some of the concepts from Deep learning that have been used to model text and images.
\subsection{Recurrent Neural Networks}
\begin{figure}[h]
    \centering
    \includegraphics[width=\textwidth]{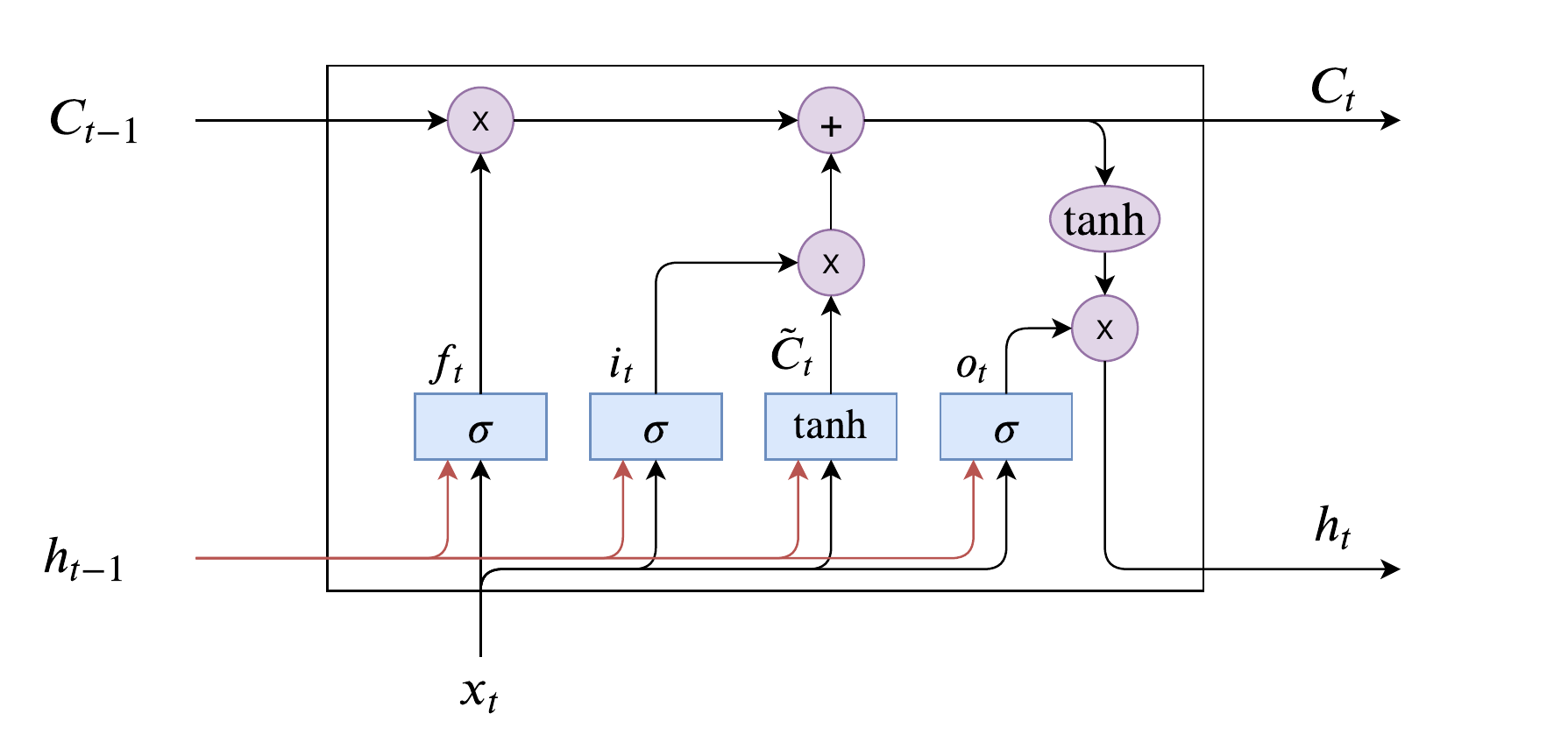}
    \caption{A diagram of the LSTM model and the interactions between the gates, the hidden state and the cell state. The blue squares represent a Fully Connected Neural Network, with the activation assigned in the blue square. The purple circles are point-wise operations and the type of operation is noted inside the circle. The diagram was inspired by the diagrams in \cite{colah}.}
    \label{fig:lstm}
\end{figure}

\begin{figure}[h]
    \centering
    \includegraphics[width=\textwidth]{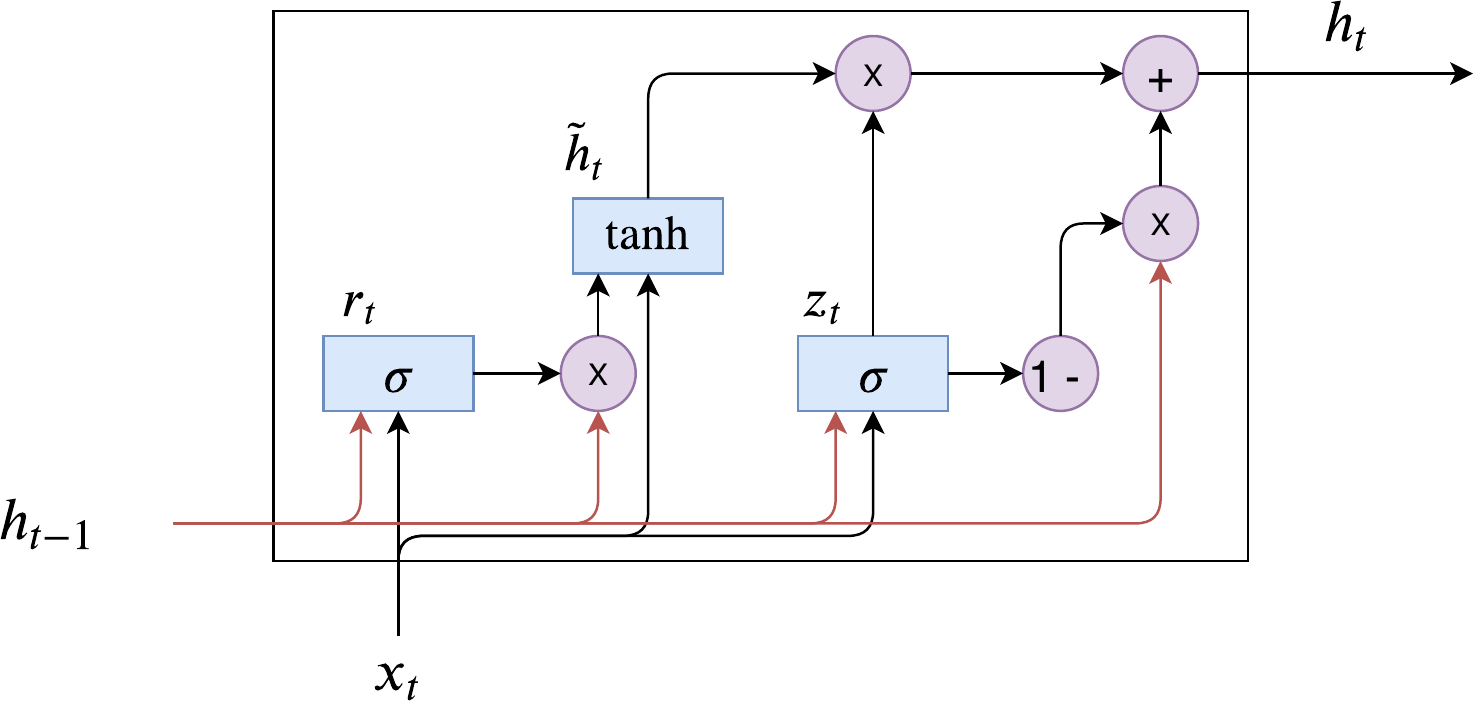}
    \caption{A diagram of the GRU model and the interactions between the gates and the hidden state. The blue squares represent a Fully Connected Neural Network, with the activation assigned in the blue square. The purple circles are point-wise operations and the type of operation is noted inside the circle. The diagram was inspired by the diagrams in \cite{colah}.}
    \label{fig:gru}
\end{figure}

Recurrent Neural Networks (RNNs) \cite{rumelhart1988learning} are neural networks designed for processing and learning the patterns in sequential data. This is achieved by including a recurrent connection to the same network layer, which enables the network to keep knowledge of the previous elements in the sequence. With this, the recurrent network can learn the dependencies between the elements of the sequence. One of the problems that RNNs face is that they are susceptible to the vanishing and exploding gradient problem \cite{hochreiter1997long} (gradients become 0 or gradients become very large numbers respectively, which results in the network not learning anything from the input). As a solution for this problem \cite{hochreiter1997long} proposed the Long Short-Term Memory (LSTM) model, which introduced the idea of representing the neural network in the RNN (cell) as a gated unit with multiple neural networks. The unit consists of a cell state (which contains the memory of all the previous inputs in the LSTM), hidden state (the output) and three gates. The forget gate controls how much of the previous cell state will affect the current cell state, the input gate controls how much the new input will affect the current hidden state and the output gate controls how the current cell state effects the output of the LSTM. Because of these gates, the LSTM has a proven ability to learn and exploit long-term dependencies in sequential data, especially in textual data \cite{sundermeyer2012lstm}. The following formulas give a better representation of how the LSTM recurrent neural network cell is constructed:
\begin{equation}
   f_t = \sigma(W_f*[h_{t-1}, x_t] + b_f) 
\end{equation}
\begin{equation}
    i_t = \sigma(W_i*[h_{t-1}, x_t] + b_i)
\end{equation}
\begin{equation}
    \widetilde{C}_t = \sigma(W_C*[h_{t-1}, x_t] + b_C)
\end{equation}
\begin{equation}
    C_t = f_t * C_{t-1} + i_t * \widetilde{C}_t
\end{equation}
\begin{equation}
    o_t = \sigma(W_o*[h_{t-1}, x_t] + b_o)
\end{equation}
\begin{equation}
    h_t = o_t * \tanh(C_t)
\end{equation}
where \(x_t\) is the input in the LSTM in current time step \(t\), \(h_{t-1}\) is the hidden state of the LSTM from the previous time step, \(h_{t}\) is the hidden state of the LSTM in the current time step, \(f_t\) is the forget gate, \(i_t\) is the input gate, \(o_t\) is the output gate, \(C_{t-1}\) is the cell state of the LSTM from the previous time step, \(\widetilde{C}_t\) is the intermediate cell state of the LSTM in the current time step  and \(C_{t}\) is the cell state of the LSTM in the current time step. \(W_f, W_i, W_C, W_o\) are weight matrices and \(b_f, b_i, b_C, b_o\) are bias vectors and together they represent the learnable parameters of the LSTM. For better visualisation of the gates and their interactions, a LSTM cell is shown in Figure \ref{fig:lstm}.
There are many variations of the LSTM network, but one that is worth mentioning is the Gated Recurrent Unit (GRU) \cite{DBLP:journals/corr/ChoMGBSB14}. It differs from the LSTM in that it combines the forget and input gates into an update gate. It also merges the cell state and hidden state, which results in a much simpler network that is easier to implement and is computationally efficient. The following formulas describe the gated unit of the GRU recurrent neural network:
\begin{equation}
   z_t = \sigma(W_z*[h_{t-1}, x_t]) 
\end{equation}
\begin{equation}
   r_t = \sigma(W_r*[h_{t-1}, x_t]) 
\end{equation}
\begin{equation}
   \tilde{h}_t = \tanh(W*[r_t*h_{t-1}, x_t]) 
\end{equation}
\begin{equation}
   h_t = (1 - z_t) * h_{t-1} + z_t * \widetilde{h}_t 
\end{equation}
where \(x_t\) is the input in the GRU cell in current time step \(t\), \(h_{t-1}\) is the hidden state of the GRU from the previous time step, \(h_{t}\) is the hidden state of the GRU in the current time step, \(z_t\) and \(r_t\) are regulating gates and \(\tilde{h}_t\) is the intermediate cell state of the GRU in the current time step. \(W_z, W_r, W\) are weight matrices and they are the learnable parameters of the GRU. For better visualisation of the gates and their interactions, the GRU cell is shown in Figure \ref{fig:gru}.
\subsection{Convolutional Neural Networks}
Deep learning has also been very successful in various vision task such as image recognition \cite{he2016deep} and image segmentation \cite{badrinarayanan2017segnet}. This success can mostly be attributed to the usage of deep convolutional neural networks. Convolutional neural networks \cite{LeCunBDHHHJ89} are neural networks that use the function convolution instead of general matrix multiplication in at least one of their layers. Convolution is a linear operation that can be explained as a sliding window function applied to parts of a matrix. The reason for their modelling power is the combination of architectural ideas such as local receptive fields, shared weights and spatial (or temporal) subsampling \cite{lecun1998gradient}. For multi-modal tasks like Image Captioning \cite{you2016image} and Visual Question Answering \cite{antol2015vqa}, researchers usually use a pre-trained deep convolutional neural network (like VGG-19 \cite{simonyan2014very} or ResNet \cite{he2016deep}), that has been trained for a task like image recognition, to obtain a vector representation of the features of every image. These vector representations are obtained by taking the output of the layers that are aligned before the 1000 dimensional classification layer. They serve as image embeddings, which means that an image of dogs would be closer to an image of a group of people, then an image of a single cat, because the image embeddings contain information about the visual context of the image which is very useful in multi-modal tasks, such as our record linkage problem. The benefit of this procedure is also computational efficiency because the image embeddings are only pre-computed and stored once, therefore making the training process of the neural network faster.
\section{The link between Visual Question Answering and Record Linkage}
All of the previously mentioned research in Record Linkage \cite{Mudgal:2018:DLE:3183713.3196926}, \cite{doi:10.1080/01621459.1969.10501049}, \cite{Bilenko:2003:ANM:1137237.1137369}, \cite{mccallum2000efficient}, \cite{hassanzadeh2009framework}, \cite{Gruenheid}, \cite{KondaDCDABLPZNP16}   has been done on textual data. The rise of digital media has provided additional information for the description of entities (such as products, advertisements etc.), which means that the above-mentioned methods have to be adjusted to take into consideration the data coming from the corresponding images. Therefore this makes record linkage a multi-modal problem since the system should combine the information that is available from the image and description of the entity and produce a joint embedding for each record. The lack of research in multi-modal record linkage forced us to explore the fields of Image Captioning and Visual Question Answering because their research is focused on modelling both textual and visual data. Our problem relates better to Visual Question Answering, where the joint representation of the question and the image is used to produce an answer, whereas in Image Captioning there is an inline representation of the caption and the image since the produced caption is conditioned on the image representation. 
\subsection{Visual Question Answering}
Visual question answering (VQA) \cite{DBLP:journals/corr/abs-1709-08203} is a task where a model needs to learn how to produce a correct answer from a given image and a question about that image. It is a unique challenge as it requires the ability to encode multi-modal input, while also being able to extract knowledge beyond a single sub-domain. For the VQA task, the algorithm needs to decide what is the relevant information, fetch that information from the image and the question, and use that to answer the questions \cite{DBLP:journals/corr/abs-1709-08203}.

The simplest architecture for VQA consists of a pre-trained convolutional neural network for image feature extraction and an LSTM for modelling the question \cite{7410636}. The output of the two networks is then concatenated and it is fed to a classifier that outputs the answer. The LSTM, due to its ability to capture dependencies between elements in a sequence has enjoyed a lot of success in the modelling language. Sentences are an ordered sequence of words, hence the LSTM is able to learn something about the underlying structure of the sentence and the dependencies between the words \cite{sundermeyer2012lstm}. Likewise, convolutional neural networks have proved to be extremely accurate in object detection and pre-trained CNN have been used in multi-modal models to increase the accuracy and decrease training time of the models. Other variations of this model use bidirectional LSTMs \cite{DBLP:journals/corr/NamHK16}, stacked LSTMs \cite{7410636} and a convolutional neural network to model the question\cite{YangHGDS15}.
\subsection{The Attention Mechanism}
To improve the standard VQA pipeline \cite{YangHGDS15} proposed a solution that implements an attention mechanism, whose goal is to find a better correlation between the encoding of question and the visual features. Attention mechanisms were introduced and successfully used in machine translation \cite{bahdanau2014neural} as a solution to the bottleneck that was the fixed-length vector produced by the recurrent neural network that was encoding the source sentence \cite{cho2014properties}. The attention mechanism allows the model to automatically (soft-)search for parts in the source sentence that are relevant to predicting a target word, without having to form these parts as a hard segment explicitly \cite{bahdanau2014neural}. The attention mechanism in machine translation produces a probability distribution over the parts of the source sentence so that the machine translation system can focus on the most probable parts of the source sentence when it is trying to produce a correct translation. Attention has been also used in image captioning \cite{DBLP:journals/corr/XuBKCCSZB15} and there the attention layer provides a probability distribution over all the pixels of the image. The result is a higher probability on the pixels of the image that are relevant to the task. In the case of \cite{YangHGDS15} the attention layer creates a query from the question encoding and uses it to obtain a probability distribution over regions of the images. They find that one attention layer is not enough, therefore their proposed solution is to use two stacked attention layers in which the second attention layer uses the output of the first attention layer as the new query which is used to produce the probability distribution over the regions of the image.
\section{Siamese Neural Networks}
As mentioned in section 2.1, another problem in record linkage is the choice of metrics that will be used to compute the similarity between the vector representations of the two records. Since the similarity metric is a function with two inputs and one output it is possible to approximate this function using a deep learning architecture known as Siamese neural network. Siamese neural networks were first introduced by \cite{bromley1994signature} to solve signature verification as an image matching problem. The Siamese network consists of two branches of networks that share the same architecture and the same set of weights. The output of the two branches is joined by an energy function at the top (which can be a similarity function like cosine distance \cite{bromley1994signature} or a neural network\cite{koch2015siamese}, \cite{zagoruyko2015learning}). Siamese networks have found usage in signature matching \cite{bromley1994signature} and recently have been used in one-shot learning for the task of image classification\cite{koch2015siamese}\cite{zagoruyko2015learning}\cite{vinyals2016matching}. For the architecture of the two branches \cite{koch2015siamese} propose multiple stacked fully connected neural network layers, \cite{bromley1994signature} propose two convolutional layers, whereas \cite{vinyals2016matching} propose a deep fully convolutional neural network. For the energy function \cite{koch2015siamese}, \cite{zagoruyko2015learning} and \cite{vinyals2016matching} propose a fully connected network that will output a single number as the similarity between the two distinct inputs. Using the Siamese neural network as part of our model, allows us to create a large neural network where we can use the ideas from Visual Question Answering to create a vector representation of an image and a description and then use the Siamese neural network to compute the similarity between a pair of multi-modal records.

\chapter{Methodology}
Based on previous research we propose a framework that is consisted of two separate units that fuse the visual and textual information of each record and a Siamese neural network that uses the output of the two fusing units to compute the similarity between the two items. We also describe two different architectures for multi-modal fusion that are inspired by Visual Question Answering models. 
\section{Data set and Preprocessing}
In this section, we describe the dataset we used to train and evaluate the models we propose in this chapter. In the previous chapter, we noted that record linkage is also referred to as database deduplication, therefore for our models, we used the Avito duplicate advertisement detection dataset \cite{avito}, that is publicly available on Kaggle. The dataset was collected from a Russian online store. The dataset consists of multiple databases that can be linked by the identification number of the advertisement and all the images that are associated with the advertisements. We use two data tables from the whole dataset. One table is a database that consists of pairs of identification numbers and a matching field that tells us whether the two items are similar (1) or dissimilar (0). The second table is a database that contains multiple fields for each advertisement. For our problem, we only used the "Description" and "Images" columns. The "Description" column contained the text that explained the advertisement and the "Images" column contained a list of image ID numbers that were associated with the particular advertisement. 

The preprocessing process was done in stages and we followed the techniques described in \cite{viksna2018sentiment}. First, we preprocessed the database that consists the information about the advertisements, then we calibrated the changes made from that database with the pairs database. The descriptions were in Russian, so the preprocessing of the text consisted of: lowercasing every character, substituting every number with 0 (because we wanted to keep information about numerical occurrences but not the original number), removing all characters that weren't in the Latin or Cyrillic alphabet, removing all double spaces and new line characters and transliterating every non-Russian word into its Cyrillic counter part \cite{sennrich2016edinburgh}. Russian is a morphologically rich language, therefore we used lemmatization to get the original form of each word (removing any suffixes or prefixes). We believe that lemmatization is better than stemming because it uses morphology and grammar to find the proper form of the word (for saw it will return see), whereas stemming is using heuristics to find the root of the word. It is not always useful, because for the words 'see' and 'saw', stemming will return 's'. The next step was to tokenizing the descriptions. We sorted the words by their frequency and decided to create a vocabulary of the 30,000 most frequent words as in \cite{bahdanau2014neural}. We assigned an integer to all the words in the vocabulary and substituted each word in the descriptions with their corresponding vocabulary index. The words that were not in the vocabulary were substituted with an "Unknown word" token. We also added a "Start of sentence" and "End of sentence" tokens to each of the descriptions. One standard practice in deep learning is to make the samples in a batch to have the same dimensions (in this way GPUs can be used more efficiently). By exploring our processed dataset we found that 90 \% of the descriptions were shorter than 100 words. Therefore all the sentences longer than 100 words were reduced to their first 100 words, whereas all the sentences that were shorter than 100 words were padded with the "Zero index" token, which is later handled within our models. 

To make our model more robust, we decided to randomly choose one image from the sequence of images to represent each advertisement. Based on the practices from Visual Question Answering we wanted to precompute the feature vectors for all the images that we were going to use. To do that we needed to preprocess them. The images were resized with its shorter side randomly sampled in [256, 480] for scale augmentation, a [224, 224] crop is randomly sampled from each image or its horizontal flip, with the per-pixel mean subtracted and we used the standard colour augmentation described in \cite{krizhevsky2012imagenet}.

We also deleted all advertisements that had corrupted images and descriptions of length 0 after the initial preprocessing. After this, we updated the pairs database by deleting all the pairs that had at least one item, not in the item database. In the end, our pairs dataset had 2,594,948 pairs of items with a class distribution: 67\% dissimilar pairs (class 0) and 33\% similar pairs (class 1). The dataset was split into a training, validation and test set with a split ratio of 80\%, 10\%, 10\%. When doing the split the samples were chosen randomly and the original distribution was maintained across the three subsets. 
\section{Baseline Model}
The Avito duplicate ads detection data set has not been featured in any research and there is no evidence that it has been used for training any neural network-based models. Here we propose a baseline model that is based on the winning solution on the Kaggle competition that has uses this dataset. We compute two hand-crafted features: Jaccard similarity coefficient between the tokenized descriptions and Euclidian distance between the image feature vectors that represent the record pair. The Jaccard similarity coefficient compares tokens from the two descriptions to see which tokens (words) are shared and which are distinct. It’s a measure of similarity for the two descriptions, with a range from 0.0 to 1.0, where 0.0 is 0\% similarity and 1.0 is 100\% similarity. The Jaccard similarity coefficient formula is given below:
\begin{equation}
    J(description_1, description_2) = \frac{|description_1 \cap description_2|}{|description_1 \cup description_2|}
\end{equation}
For every image, we will extract features from a pre-trained ResNet convolutional neural network \cite{HeZRS15} and then for each pair of records, we will compute the Euclidian distance between the embedded representations of the associated images. Below is the formula for Euclidean distance:
\begin{equation}
    d(img\_embed_1, img\_embed_2) = \sqrt{\sum_{i=1}^{n} (img\_embed_1^i - img\_embed_2^i)^2}
\end{equation}
There are more hand-crafted features that the winners of the competition used, but for a better comparison of the models, we only use the descriptions and images of the advertisements. We normalize both of the features in the training set using z-score normalization and we apply the same z-score normalization on the test set. Z-score normalization is done by subtracting the mean of the feature from each sample and then dividing this with the standard deviation of the feature. With this we are centering the features around their corresponding means and we also ensure that both features will be in the same range, which will help our model to generalize better. These two features are the input to a logistic regression model. 
\section{Siamese Neural Network Framework}
\begin{figure}[h]
    \centering
    \includegraphics[width=\textwidth]{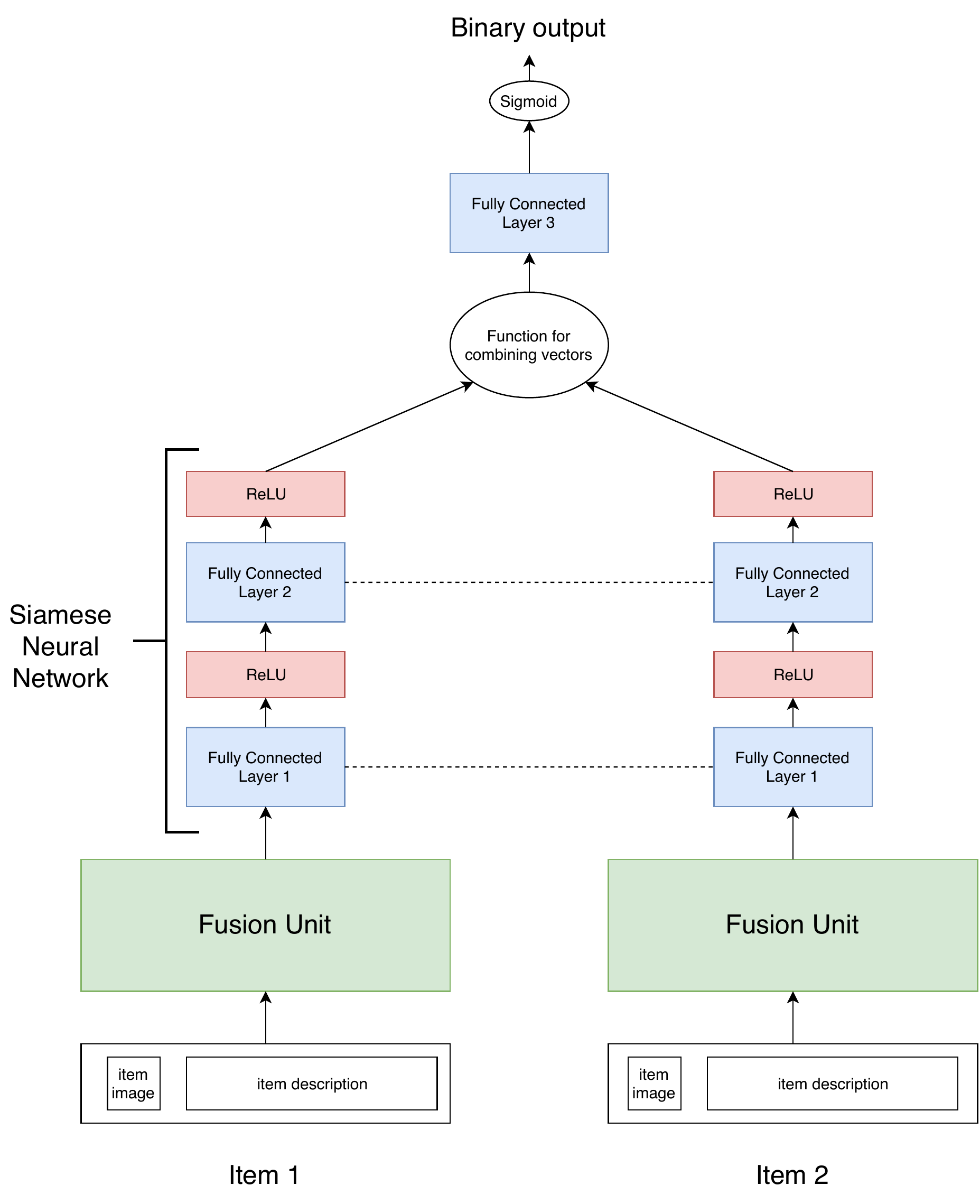}
    \caption{An illustration of our proposed framework. The fusion models are independent of the framework. The dotted lines mean that the neural network layers share the same weights. The function that combines the two output vectors is the absolute difference of the two vectors.}
    \label{fig:framework}
\end{figure}

The framework that we propose consists of a fusion unit that needs to learn how to combine the image and the description of each item in the pair. The fusion units output two vector representations each for the corresponding advertisement. These outputs are fed into a Siamese Neural network that produces a binary output. The scope of our research is the fusion models and their architecture will be explained in the sections below. We based our Siamese Neural Network model on the work of  \cite{koch2015siamese} and \cite{zagoruyko2015learning}. The model consists of two branches and an energy function on the top. The architectures of the branches are identical and share weights. The architecture of one branch consists of two fully connected neural network layers with a ReLU activation function after each layer. The outputs of the two branches are then combined by computing the absolute distance between the two outputs. After we have combined the outputs of the two branches into one vector, we feed this vector into a fully connected layer that is followed by a sigmoid function that finally produces a binary output. Figure \ref{fig:framework} illustrates the architecture of the framework.
\section{Recurrent Neural Network + Convolutional Neural Network fusion module}
\begin{figure}[h]
    \centering
    \includegraphics[width=\textwidth]{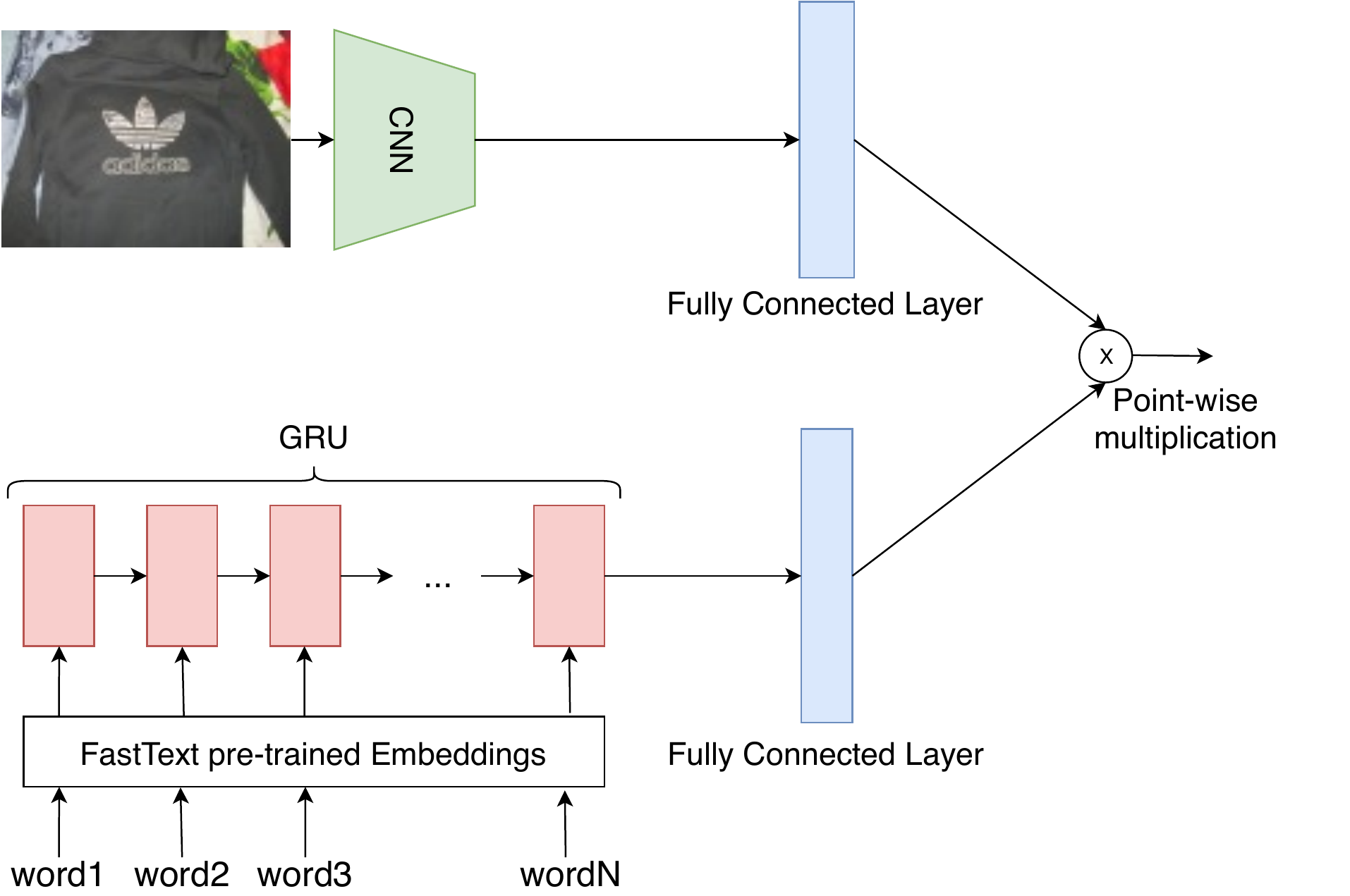}
    \caption{An illustration of the simple fusion model. The GRU is used to create a description encoding and a pre-trained ResNet to extract a vector representation of the image features. After we use fully connected layers to transform the visual and textual encoding to the same vector space, we combine the visual and textual information by point-wise multiplication of the two encodings.}
    \label{fig:standard}
\end{figure}

The first fusion module that we proposed is inspired by the architecture proposed in \cite{antol2015vqa}. The model is composed of a pre-trained ResNet152 convolutional neural network that we use to extract features from the image and a GRU recurrent neural network to model the description. We use the second to last layer of the pre-trained ResNet152, which outputs a 2048-dimensional feature vector. We decided to use GRUs instead of LSTMs because they are more computationally efficient and it has been shown that they achieve similar performance when modelling natural language \cite{bahdanau2014neural}. To encode the description, first, we will replace every token (word) with its corresponding 300-dimensional FastText word embedding \cite{bojanowski2016enriching}. We do this because the pre-trained embeddings give a better semantic vector space representation of the language than one-hot encodings\cite{Pennington14glove}. After we have applied this transformation to the description, we will feed the description to a GRU recurrent neural network. The output of the GRU represents an encoded vector representation of the description. Because the size of the description encoding and the image feature vector is different, we use two separate fully connected neural networks to transform the description encoding and the image encoding into similar sized vectors. To combine the textual and visual information, we use point-wise multiplication (as suggested in \cite{antol2015vqa}) to fuse the description encoding and the image feature vector, which is the final output of the fusion module. This architecture is illustrated in Figure \ref{fig:standard}.

\section{Stacked Attention Network fusion module}
\begin{figure}[h]
    \centering
    \includegraphics[width=\textwidth]{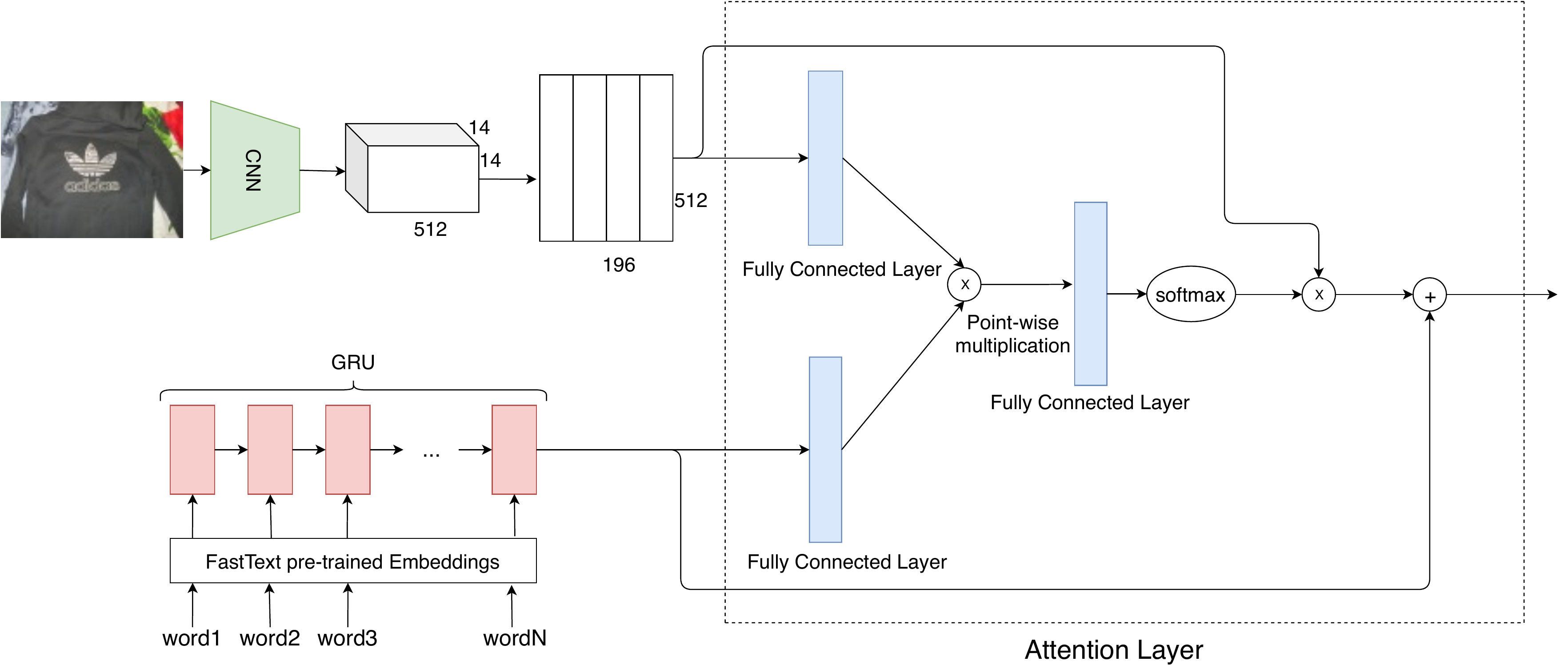}
    \caption{An illustration of the Stacked Neural Network fusion module. The GRU is used to create a description encoding and a pre-trained VGG-19 to extract a vector representation of the image features.}
    \label{fig:san}
\end{figure}

The second fusion module is inspired by the Stacked Attention Networks \cite{YangHGDS15} architecture used for Visual Question Answering. We propose that the fusion model should use a pre-trained VGG-19 convolutional neural network to obtain a feature vector representation of the image. For each image, we obtain a [14, 14, 512] vector, where 14 x 14 is the number of regions that we split our image into. Therefore we transform the output of the VGG-19 network into a [196, 512] matrix, where every 512-dimensional column is a feature representation of each region in the image. We choose VGG-19 instead of ResNet, because ResNet doesn't have an intermediate output that corresponds to a feature representation of each region. We use a GRU recurrent neural network to obtain an encoding for the description in the same way as the previously described model. For modelling convenience, we use a single fully connected neural network to transform each feature vector to a new vector that has the same dimension as the description vector. We want to find a better correlation between the description and the visual features because if two records represent the same item but the description is worded differently, we would want the vector representations of the two records to be closer. A possible solution would be to use an attention layer that will connect the description and the feature vector of the image. The attention mechanism should learn to give a higher probability mass to the feature vector of the region in the image that is relevant to the words in the description. In this module, we experiment with multiple attention layers because \cite{YangHGDS15} concluded that for complicated questions/descriptions, a single attention layer is not sufficient to locate the correct region for answer prediction if in the image there are some subtle relationships among multiple objects \cite{YangHGDS15}.
Given the image feature matrix \(v_i\) and the description encoding \(v_d\), we combine them by using a fully connected neural network and then we use softmax to produce a probability distribution (attention weights) over the regions of the image. We then calculate a weighted sum of the image feature columns using the attention distribution. Lastly, we combine the weighted sum with the description encoding to form a refined query vector. The final vector is refined because it encodes both the description information and the visual information that is relevant to the description. We implement our solution so that we can experiment with multiple numbers of attention layers. The following formula defines the \(k\)-th attention layer and gives a mathematical representation of the description above:
\begin{equation}
    h^{k}_{A} = \tanh(W^{k}_{I,A}v_{I} \oplus (W^{k}_{Q,A}u^{k-1} + b^{k}_{A}))
\end{equation}
\begin{equation}
    p^{k}_{I} = \textbf{softmax}(W^{k}_{P}h^{k}_{A}+b^{k}_{P})
\end{equation}
\begin{equation}
    \tilde{v}^{k}_{I} = \sum_{i} p^{k}_{i}v_{i}
\end{equation}
\begin{equation}
    u^{k}=\tilde{v}^{k}_{I}+u^{k-1}
\end{equation}
where \(u^{0}\) is initialized to be the initial description embedding \(v_d\), \(v_i\) are the image features, \(h^{k}_{A}\) is the intermediate attention layer, \(p^{k}_{I}\) is the probability distribution over each region, \(\tilde{v}^{k}_{I}\) is the weighted sum of the attention distribution and the image features and \(u^{k}\) is the new query. \(W^{k}_{I,A}, W^{k}_{Q,A}, W^{k}\) are weight matrices and \(b^{k}_{A}, b^{k}_{P}\) are biases and together they are the learnable parameteres of the attention mechanism.
This is repeated for \(K\) times and then the final \(u^K\) will be used as one of the inputs to a Siamese neural network. Same as the second model, to train the Siamese neural network, for each item in the pair of similar or dissimilar records, we will need to obtain \(u^K\) for both entities. In Figure \ref{fig:san} we illustrate a single attention layer SAN network module. 
\chapter{Results and Evaluation}
\section{Evaluation Metric}
Before we present all the results that we gathered from our experiments we will explain our evaluation metric. Record linkage is a binary classification problem and we decided to use Average Precision Score to evaluate our models. In the following section, we will refer to class 1 as the positive class (the pair of items is similar) and class 0 as the negative class (the pair of items is not similar). Average Precision Score summarizes the area under the Precision-Recall curve as the weighted mean of precisions achieved at different thresholds, with the increase in recall from the previous threshold used as the weight \cite{aps}:
\begin{equation}\label{eq:4.1}
    \text{AP} = \sum_n (R_n - R_{n-1}) P_n
\end{equation}
where \(P_n\) and \(R_n\) are the precision and recall at the \(n\)-th threshold. Precision and Recall gives us information about the ratios of true positives \textit{TP}, false positives \textit{FP} and false negatives \textit{FN}. \textit{TP} is the number of examples that were positive and that the model classified as positive, \textbf{FP} is the number of examples that were negative but the model predicted that they were positive and \textit{FN} is the number of examples that were positive but the model predicted that they were negative. Precision is calculated as:
\begin{equation}
    \text{P} = \frac{\textit{TP}}{(\textit{TP} + \textit{FP})}
\end{equation}
It describes how good the model is at predicting the positive class, which in our case is class 1, the less represented class. Recall is calculated as:
\begin{equation}
    \text{R} = \frac{\textit{TP}}{(\textit{TP} + \textit{FN})}
\end{equation}
Recall measures the proportion of true positives that are correctly identified and it is also referred to as sensitivity. Precision usually refers to precision at a certain threshold. For example, if we classify any output of the model that is less then 0.5 as negative and positive if the output is greater then 0.5, then the precision score is accurate only for threshold = 0.5. If we were to change the threshold, the precision score would change. When the class distribution is unbalanced (as in our case) we would like to vary this threshold, to get a better understanding of how the model is performing. Average Precision Score gives us the average precision at all such possible thresholds, without having to specify the decision thresholds. This implementation \ref{eq:4.1} is not interpolated and is different from computing the area under the precision-recall curve with the trapezoidal rule, which uses linear interpolation and can be too optimistic \cite{aps}. Even though we choose the models based on the Average Precision Score, we compare the Precision-Recall curves for each of the models we propose, to get a better visualization of the performance of our model. The Precision-Recall curve shows the tradeoff between precision and recall at different thresholds. A large area under the curve means that both recall and precision are high,  where high precision relates to a low \textbf{FP} rate, and high recall relates to low \textbf{FN} rate. High scores for both show that the classifier is returning accurate results (high precision), as well as returning a majority of all positive results (high recall) \cite{aps}.

\section{Baseline Model}
\begin{figure}
\centering
    \begin{minipage}{.5\textwidth}
      \centering
      \includegraphics[width=\linewidth]{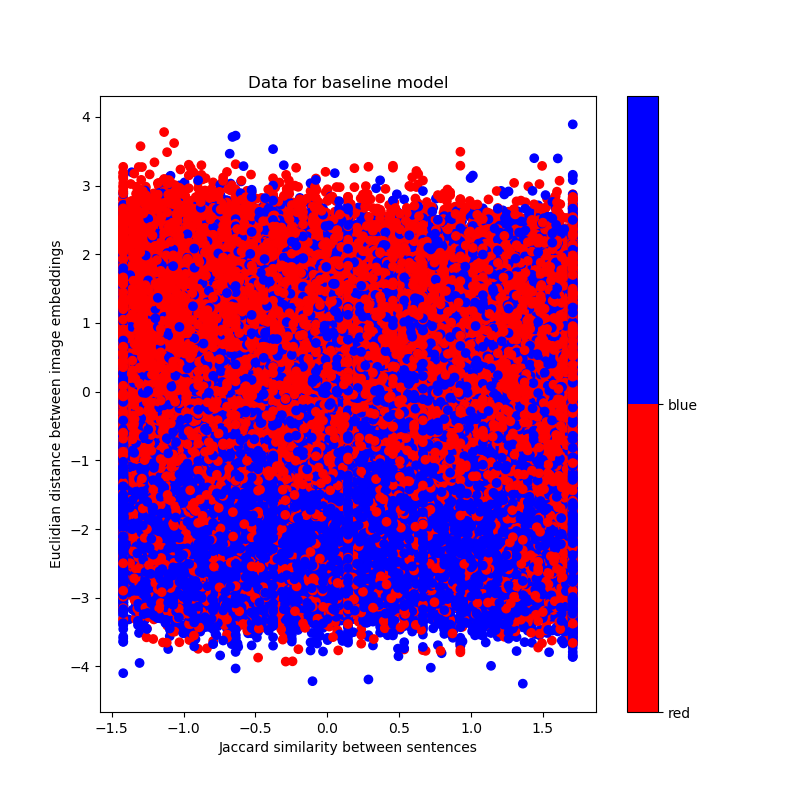}
      \captionof{figure}{Actual dist. of training set.}
      \label{fig:baseline_train_data}
    \end{minipage}%
    \begin{minipage}{.5\textwidth}
      \centering
      \includegraphics[width=\linewidth]{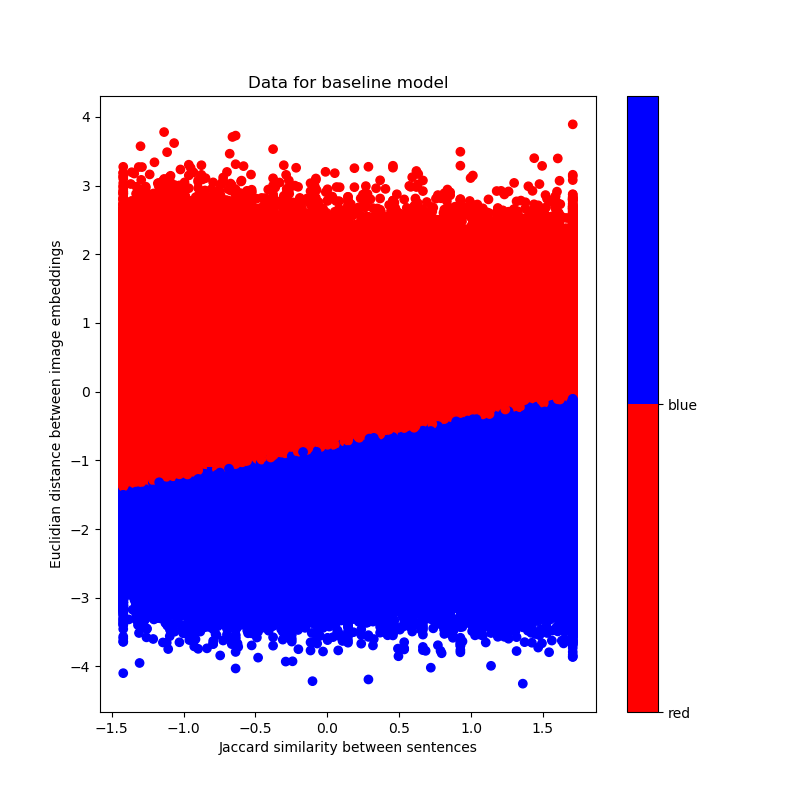}
      \captionof{figure}{Predicted dist. of training set.}
      \label{fig:baseline_train_pred}
    \end{minipage}
    \centering
    \begin{minipage}{.5\textwidth}
      \centering
      \includegraphics[width=\linewidth]{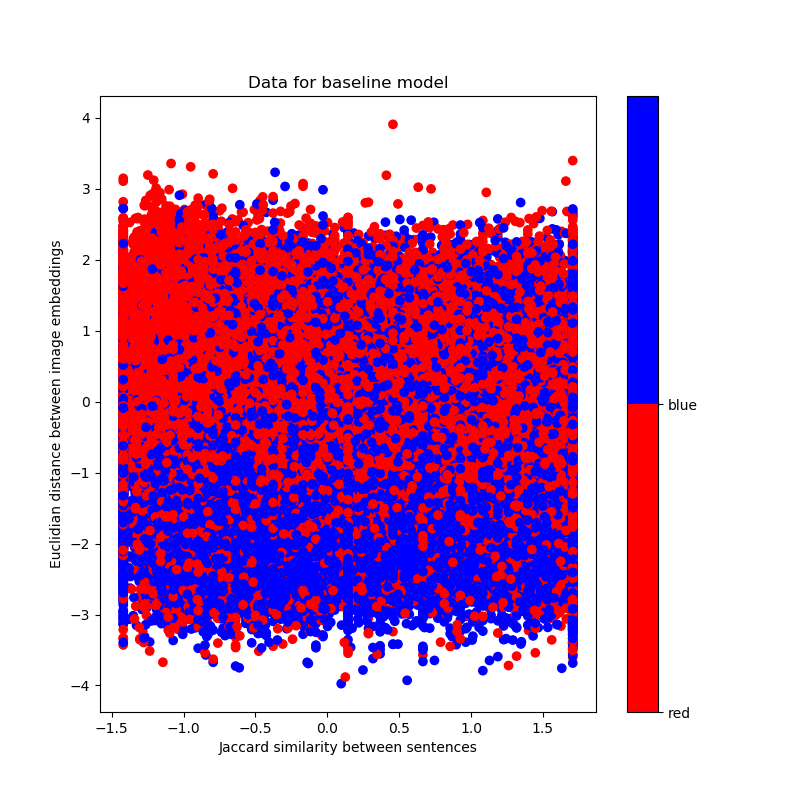}
      \captionof{figure}{Actual dist. of test set.}
      \label{fig:baseline_test_data}
    \end{minipage}%
    \begin{minipage}{.5\textwidth}
      \centering
      \includegraphics[width=\linewidth]{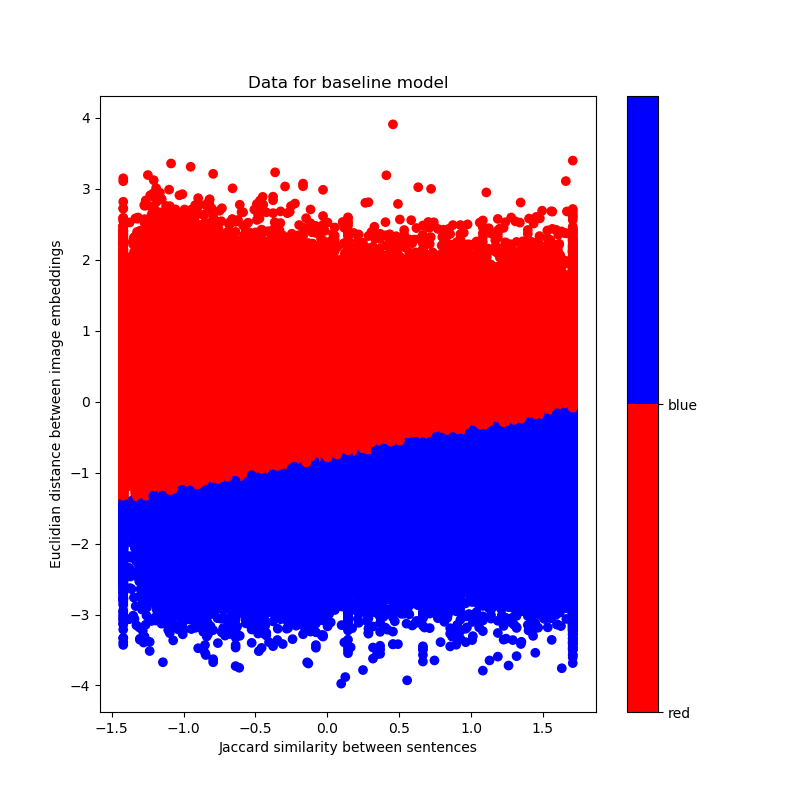}
      \captionof{figure}{Predicted dist. of test set.}
      \label{fig:baseline_test_pred}
    \end{minipage}
    \caption{In the plots, the red dots represent samples from the negative class and the blue dots are samples from the positive class.}
\end{figure}

\begin{figure}
    \centering
    \includegraphics[width=\textwidth]{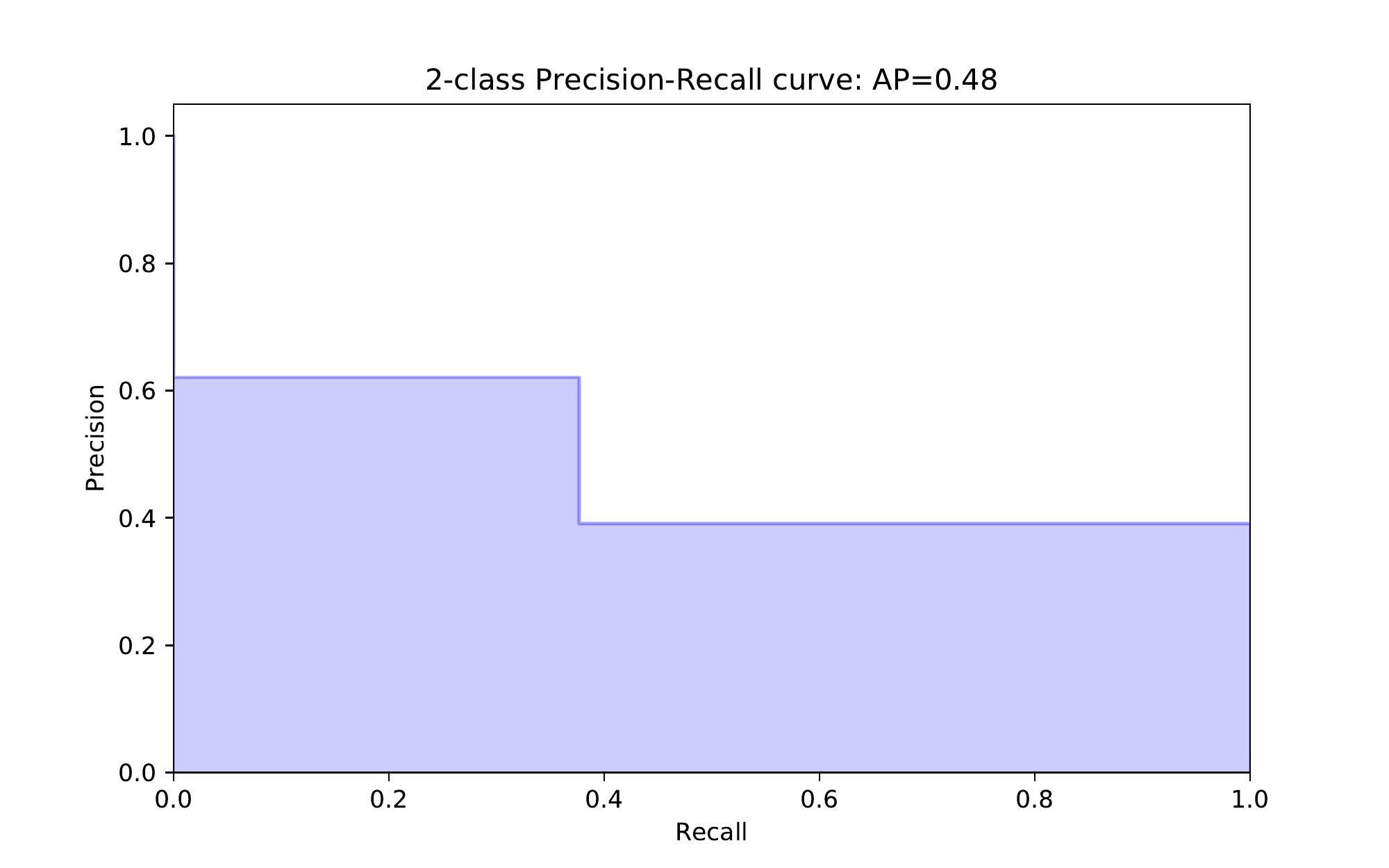}
    \caption{Precision - Recall curve produced by the baseline model on the training set.}
    \label{fig:PR-Curve_train}
\end{figure}

\begin{figure}
    \centering
    \includegraphics[width=\textwidth]{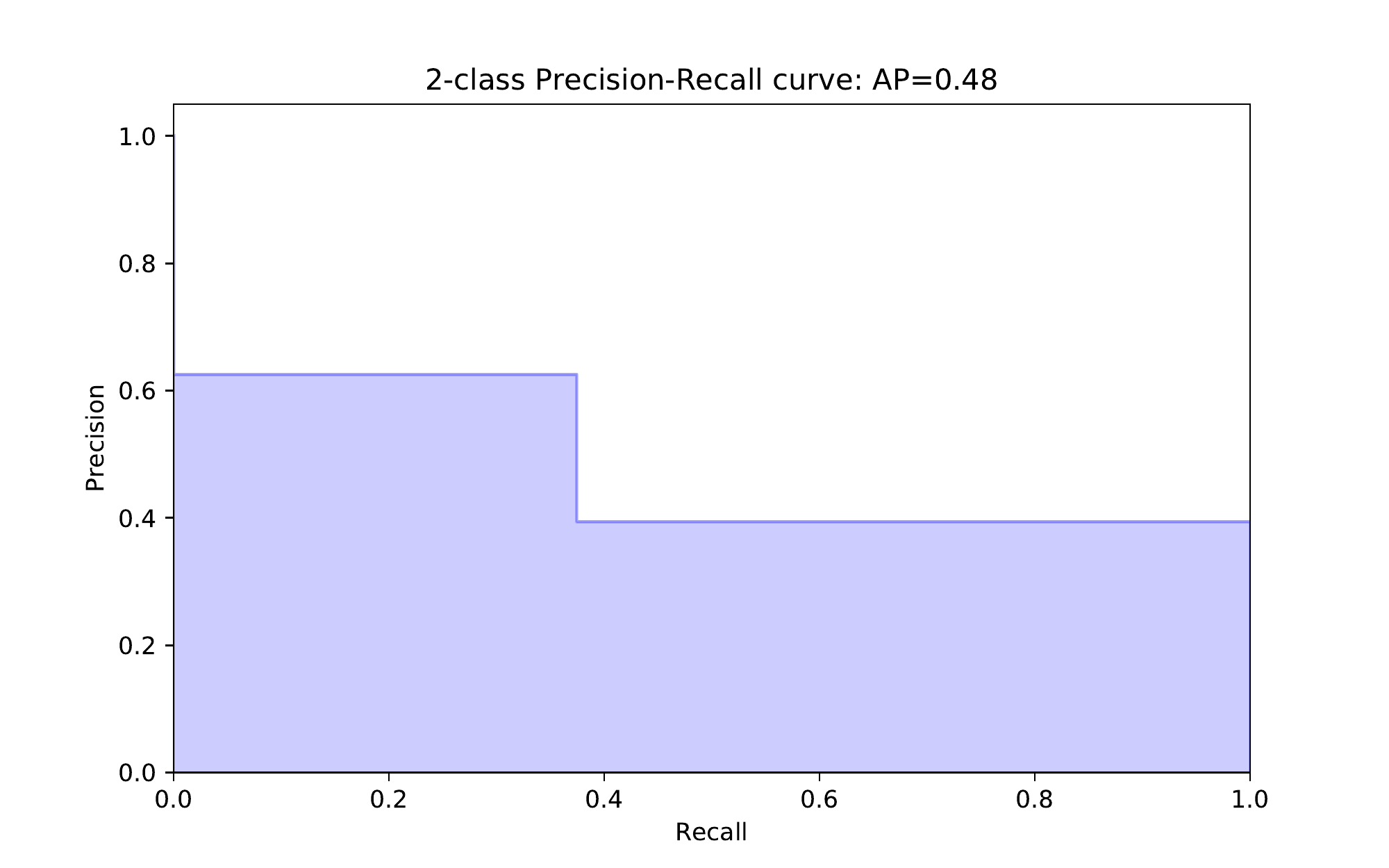}
    \caption{Precision - Recall curve produced by the baseline model on the test set.}
    \label{fig:PR-Curve_test}
\end{figure}

The baseline model is a logistic regression model trained on a training data set with two hand-crafted features (Euclidian distance between the image feature vectors of advertisements and Jaccard similarity coefficient between the descriptions of the advertisements in the pair). The baseline model achieved an average precision score of \textbf{0.4776} on the \textbf{training set} and an average precision score of \textbf{0.4808} on the \textbf{test set}. Figures \ref{fig:baseline_train_data} and \ref{fig:baseline_test_data} illustrates the actual distribution of the training set and test set respectively, while in Figures \ref{fig:baseline_train_pred} and \ref{fig:baseline_test_pred} we can see how the model predicts the distributions of the training and test set. From the figures it is obvious that the logistic regression model has learned the coefficients of a linear function that minimizes the negative-log-likelihood loss. Figures \ref{fig:PR-Curve_train} and \ref{fig:PR-Curve_test} are almost identical and they illustrate the Precision - Recall curve for the baseline model on the training and test set. The figures show that the baseline model has a precision above 0.6 when the recall is in the range of 0.0 and 0.4. For bigger recall the precision drops to 0.4. This means that when the baseline model produces more positive outcomes, it predicts the positive examples more accurately but it also predicts a large number of negative examples as positive. 

\section{Training the Neural Network model}
The neural network based solutions that we propose are based on two fusion modules that combine the textual and visual information of each advertisement and a Siamese Neural Network that takes the output of the fusion modules as its input and produces an output that is a number between 0 and 1. The output of this model is the probability that the two advertisements are similar, therefore if the output is closer to 1 it means that the model predicts with high certainty that the advertisements are similar. Neural networks are trained by updating their weights using Backpropagation (as explained in Section 2.2.1) and for that we need to compute the difference between the output of the model and the actual value. This is done by using a differentiable loss function. The labels in our dataset are either 0's or 1's and our model outputs probabilities that are in the range of 0 to 1. This means that we need to penalize the model when it outputs a high probability but the true label is 0 or it outputs a low probability but the actual label is 1. This can be done with the Binary Cross Entropy loss function, that is defined as:
\begin{equation}
    BCE(x, y) =  - (y*\log{p(x)} + (1-y)*\log{(1-p(x))})
\end{equation}
where \(x\) is the pair of advertisements and the input to the model, \(y\) is the label and \(p(x)\) is the output of our model. By minimizing the Binary Cross-Entropy loss function we will force the model to learn how to adjust its weights, so when it produces a probability about the similarity of the pair of advertisements, it is as close as possible to the true label. Our choice of the loss function is also supported by \cite{koch2015siamese} and \cite{zagoruyko2015learning}, who also use Binary Cross Entropy to compute the loss between the true label and the output of their Siamese neural networks. It is also worth noting that Binary Cross Entropy is just an approximation of the Average Precision Score metric. Decreasing the BCE loss doesn't have to correlate to an increase in the Average Precision Score, but we assume that is the case. We have to use Binary Cross Entropy instead of Average Precision Score because APS is not a differentiable function. We will use BCE to train the model, but we also compute the Average Precision Score after every epoch in the Stochastic Gradient Process, because that is the metric that we use to compare models.
\section{Recurrent Neural Network + Convolutional Neural Network fusion module}
The first proposed fusion module consists of a GRU that encodes the description, a pre-trained ResNet that outputs a vector representation of the image and two fully connected layers that transform the image feature representation and description encoding to the same vector space. The outputs of the two fully connected layers are combined by point-wise multiplication. To find the best hyperparameter configuration for the fusion module and Siamese neural network we did a small hyperparameter search by training 10 different models. We explored the number of stacked GRU layers (layers), the size of the GRU cell (hid\_size), the size of the two fully connected layers (fs\_out\_size), the size of the first fully connected layer in the Siamese neural network (fc1\_hid\_size), the size of the second fully connected layer in the Siamese neural network (fc2\_hid\_size) and the learning rate (lr.). All models were trained for 30 epochs and as an optimization technique, we used early stopping (we stopped the training loop if after 5 epochs the validation loss didn't improve). Due to the exponential complexity of doing a more in-depth hyperparameter search, we decided that we are going to train our models with mini-batches of 64 samples and use a fixed random seed so that the weights in the neural networks in our models would always have the same random initialization. By doing this we would be able to compare the results of the models more accurately. The results of the hyperparameter search are shown in Table \ref{ref:standard_train_table} and Table \ref{ref:standard_val_table}. Table \ref{ref:standard_train_table} shows the Binary Cross Entropy loss and the Average Precision Score. the model achieved on the training set, while Table \ref{ref:standard_val_table} show the same metrics the model achieved on the validation set. We are more interested in the scores the model achieves on the validation set, because the model hasn't seen the validation data (hasn't been trained on it), therefore the scores achieved on the validation set give us a more honest outlook on how the model generalizes and how it will perform in the future on unseen data from the real world. 

\begin{table}[h!]
\centering
 \begin{tabular}{|c|c|c|c|c|c|c|c|} 
 \hline
 layers & hid\_size & fs\_out\_size & fc1\_hid\_size & fc2\_hid\_size & lr. & train\_loss & train\_aps \\ [0.5ex] 
 \hline
 1 & 100 & 100 & 64 & 32 & 0.1 & 15.23 & 0.391 \\ [0.5ex] 
 \hline
 1 & 100 & 100 & 64 & 32 & 0.01 & 13.45 & 0.392 \\ [0.5ex] 
 \hline
 1& 100 & 100 & 64 & 32 & 0.001 & 10.20 & 0.387 \\ [0.5ex] 
 \hline
 1& 100 & 100 & 64 & 32 & 0.0001 & 0.5241 & 0.741 \\ [0.5ex] 
 \hline
 1& 256 & 256 & 128 & 64 & 0.1 & 14.05 & 0.393 \\ [0.5ex] 
 \hline
 1& 256 & 256 & 128 & 64 & 0.01 & 11.43 & 0.398 \\ [0.5ex] 
 \hline
 1& 256 & 256 & 128 & 64 & 0.001 & 10.15 & 0.413 \\ [0.5ex] 
 \hline
 1& 256 & 256 & 128 & 64 & 0.0001 & \cellcolor{red!25}0.463 & \cellcolor{red!25}0.802 \\ [0.5ex] 
 \hline
 2& 512 & 1024 & 512 & 1024 & 0.01 & 14.94 & 0.3912 \\ [0.5ex] 
 \hline
 2& 512 & 1024 & 512 & 1024 & 0.0001 & 6.67 & 0.454 \\ [0.5ex] 
 \hline
 \end{tabular}
 \caption{Results the models achieved on the training set. The highlighted cells are the lowest training loss and highest training average precision score that were achieved by the models that we trained.}
 \label{ref:standard_train_table}
\end{table}

\begin{table}[h!]
\centering
 \begin{tabular}{|c|c|c|c|c|c|c|c|} 
 \hline
 layers & hid\_size & fs\_out\_size & fc1\_hid\_size & fc2\_hid\_size & lr. & val\_loss & val\_aps \\ [0.5ex] 
 \hline
 1 & 100 & 100 & 64 & 32 & 0.1 & 10.83 & 0.392 \\ [0.5ex] 
 \hline
 1 & 100 & 100 & 64 & 32 & 0.01 & 16.79 & 0.392 \\ [0.5ex] 
 \hline
 1& 100 & 100 & 64 & 32 & 0.001 & 7.56 & 0.375 \\ [0.5ex] 
 \hline
 1& 100 & 100 & 64 & 32 & 0.0001 & 0.581 & 0.668 \\ [0.5ex] 
 \hline
 1& 256 & 256 & 128 & 64 & 0.1 & 7.74 & 0.388 \\ [0.5ex] 
 \hline
 1& 256 & 256 & 128 & 64 & 0.01 & 11.12 & 0.401 \\ [0.5ex] 
 \hline
 1& 256 & 256 & 128 & 64 & 0.001 & 8.76 & 0.44 \\ [0.5ex] 
 \hline
 1& 256 & 256 & 128 & 64 & 0.0001 & \cellcolor{red!25}0.587 & \cellcolor{red!25}0.684 \\ [0.5ex] 
 \hline
 2& 512 & 1024 & 512 & 1024 & 0.01 & 10.83 & 0.392 \\ [0.5ex] 
 \hline
 2& 512 & 1024 & 512 & 1024 & 0.0001 & 0.653  & 0.516 \\ [0.5ex] 
 \hline
 \end{tabular}
 \caption{Results the models achieved on the validation set. The highlighted cells are the lowest validation loss and highest validation average precision score that were achieved by the models that we trained.}
 \label{ref:standard_val_table}
\end{table}
From the tables we can see that the model with the 8th configuration has achieved the best training and validation average precision score of \textbf{0.802} and \textbf{0.684}. The second best model is the 4th model and it is twice as smaller in size then the 8th model, but it achieved a validation average precision score of \textbf{0.668} which isn't so far from the best score. Given the results, we can conclude that the models are performing better with a small learning rate, which in our case is 0.0001. The last two models in the tables have the same setting as the best model in \cite{antol2015vqa} (for the fusion module) and for the Siamese Neural Network we used the settings from \cite{koch2015siamese}. An interesting observation from this is that the hyperparameters that were the best setting for the networks in \cite{antol2015vqa} and \cite{koch2015siamese}, didn't achieve good results for our task, which means that hyperparameters settings from papers don't always transfer well from one task to another. 
\begin{figure}
    \centering
    \includegraphics[width=\textwidth]{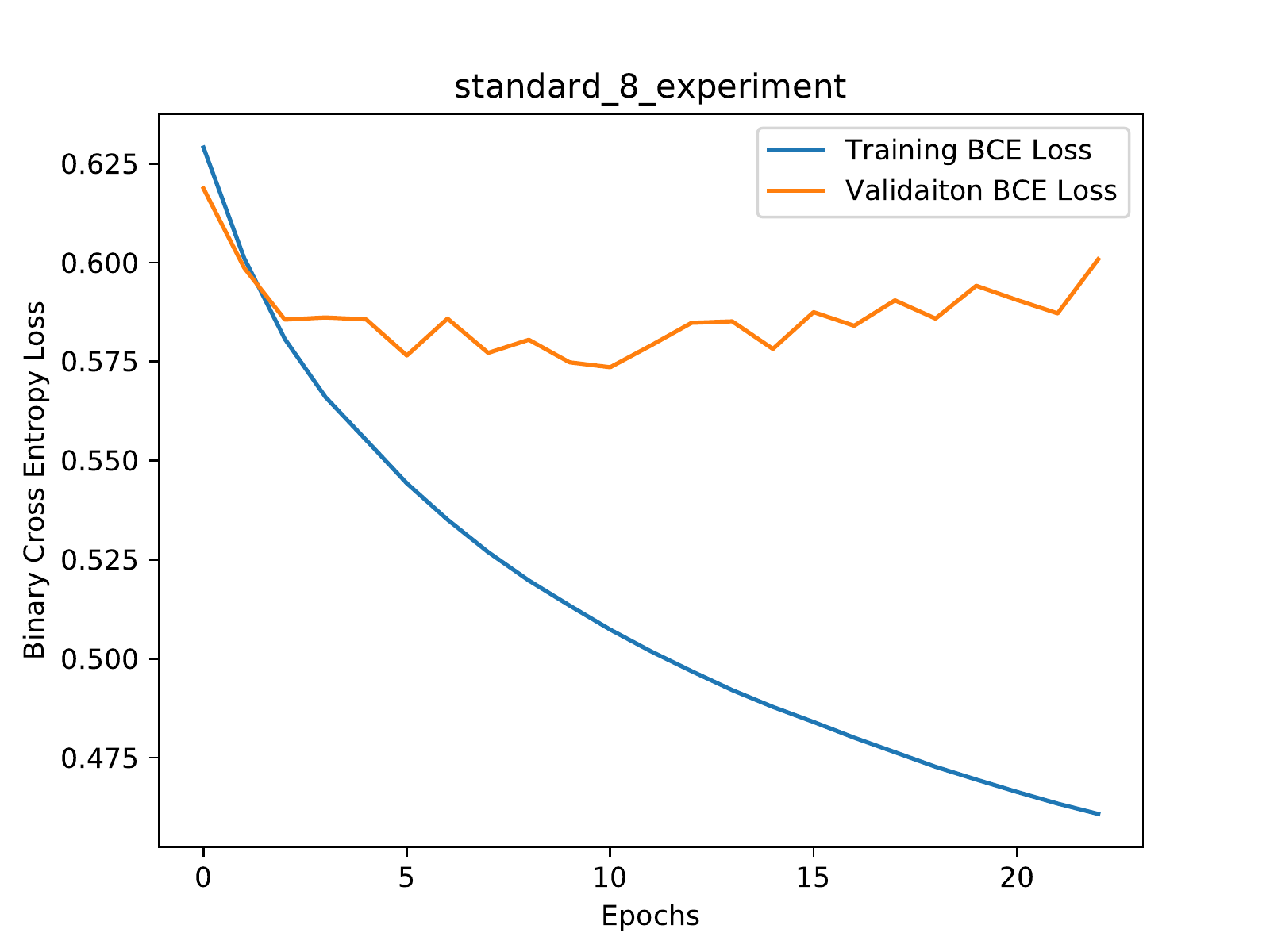}
    \caption{The change of Binary Cross Entropy loss during training. The blue line is the training loss and the orange line is the validation loss.}
    \label{fig:s8bce}
\end{figure}
\begin{figure}
 \centering
  \includegraphics[width=\textwidth]{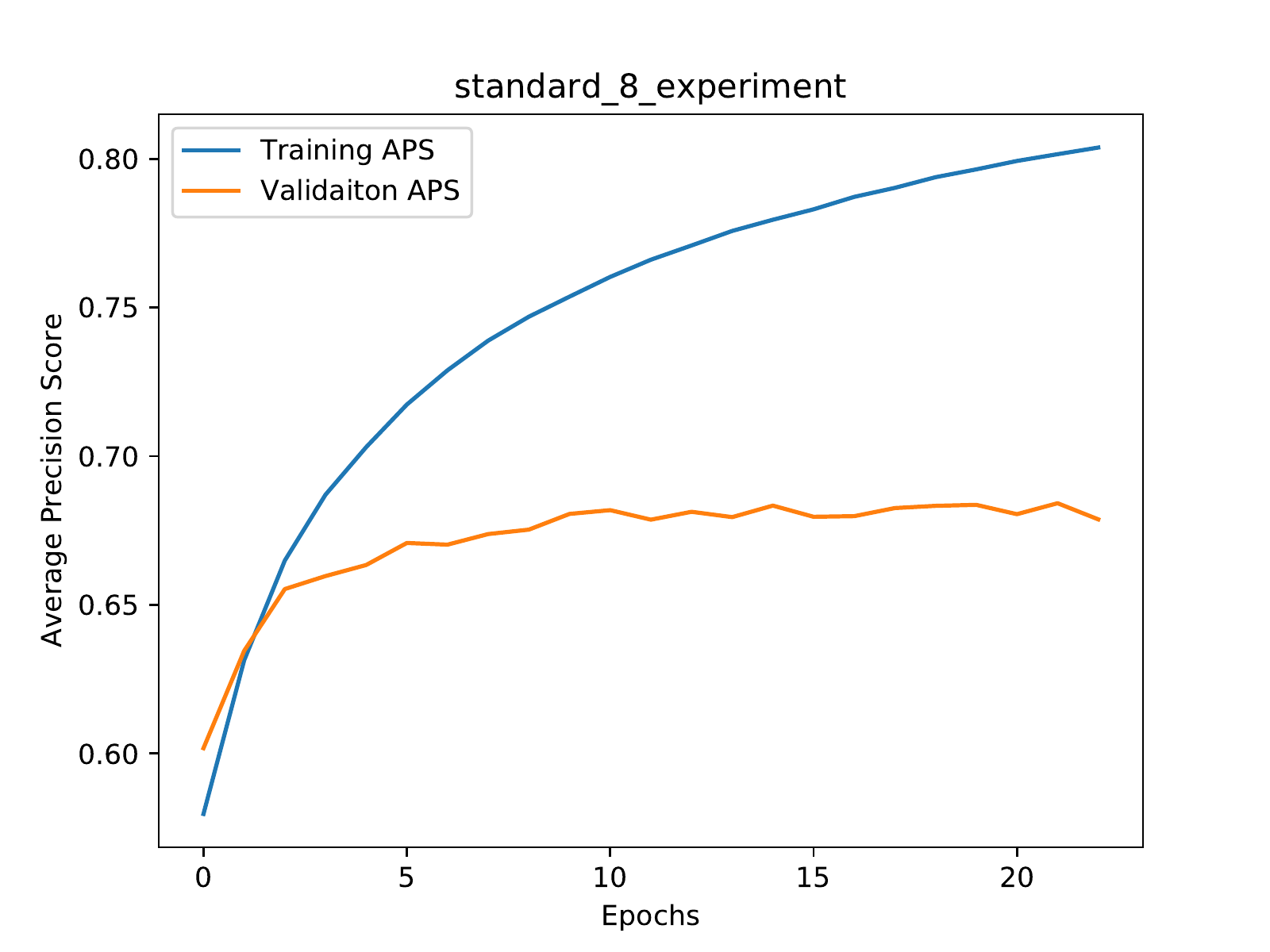}
  \caption{The change of Average Precision Score during training. The blue line is the training score and the orange line is the validation score.}
  \label{fig:s8aps}
\end{figure}     

In Figure \ref{fig:s8bce} we can see how the Binary Cross Entropy loss changes during training time and in Figure \ref{fig:s8aps} illustrates the same thing about the Average Precision Score. From the figures, we can see that after the 12th epoch the model decreases the training loss, but the validation loss seems to get bigger. This is a sign of overfitting, which means that while the model is improving his performance on the training data, it starts to lose its generalization power and performs worse on the validation set. As a solution to this, we proposed to train our model using the Dropout regularization technique \cite{srivastava2014dropout}. Dropout is a simple process that randomly sets units to be zero (this results in dropping their connections) from the neural network during training and this prevents the units from co-adapting so much. Dropout has one parameter which is the dropout rate or in other words the proportion of connection we going to forget from the input. In our model, we use Dropout right after the point-wise multiplication. By doing this we are regularizing the way the fusion module works and don't interfere with how the Siamese Neural Network learns because our main focus is to explore which fusion module works best.

For the following experiments, we explored three different dropout rates (0.2, 0.3 and 0.5) and fixed the learning rate to 0.0001. The results of the experiments can be seen in Tables \ref{ref:s8drtable} and \ref{ref:vals8drtable}.
\begin{table}[h!]
\centering
 \begin{tabular}{|c|c|c|c|c|c|c|c|} 
 \hline
 layers & hid\_size & fs\_out\_size & fc1\_hid\_size & fc2\_hid\_size & dr & train\_loss & train\_aps \\ [0.5ex] 
 \hline
 1& 256 & 256 & 128 & 64 & 0.2 & 0.591 & 0.647\\ [0.5ex] 
 \hline
 1& 256 & 256 & 128 & 64 & 0.3 & 0.6 & 0.63\\ [0.5ex] 
 \hline
 1& 256 & 256 & 128 & 64 & 0.5 & 0.624 & 0.587\\ [0.5ex] 
 \hline
 \end{tabular}
 \caption{Results the models achieved on the training set.}
 \label{ref:s8drtable}
\end{table}

\begin{table}[h!]
\centering
 \begin{tabular}{|c|c|c|c|c|c|c|c|} 
 \hline
 layers & hid\_size & fs\_out\_size & fc1\_hid\_size & fc2\_hid\_size & dr & val\_loss & val\_aps \\ [0.5ex] 
 \hline
 1& 256 & 256 & 128 & 64 & 0.2 & 0.586 & 0.654\\ [0.5ex] 
 \hline
 1& 256 & 256 & 128 & 64 & 0.3 & 0.593 & 0.643\\ [0.5ex] 
 \hline
 1& 256 & 256 & 128 & 64 & 0.5 & 0.617 & 0.605\\ [0.5ex] 
 \hline
 \end{tabular}
 \caption{Results the models achieved on the validation set.}
 \label{ref:vals8drtable}
\end{table}
From the tables, we can see that applying dropout right after the point-wise multiplication didn't improve the model's performance. Therefore in this section, we conclude that the best model from these experiments is the model with the 8th configuration from Table \ref{ref:standard_train_table}. To get a better estimation of the average precision score on the validation set, we trained two models that had the 8th configuration but we used a different random initialization of the weights of the model. After we trained the models we averaged their validation scores, therefore the final average precision score on the validation set is $\textbf{0.682} \pm 0.003$.
\section{Stacked Attention Network fusion module}
The Stacked Attention Network fusion module is an extension to the previously proposed fusion module. Instead of combining the outputs of the GRU and the pre-trained convolutional network by point-wise multiplication, SAN uses the attention mechanism to create a more refined joint representation of the description and the image features. Because of time constraints (one epoch lasted 48 hours) we were able to train only one model with this fusion module. Since the 8th configuration from Table \ref{ref:standard_val_table} performed the best in previous experiments, we decided that our model is going to have the same configuration. The only difference is that the attention layers had the same size as the output of the first fusion module.  

The average precision score that this model achieves on the validation set is \textbf{0.534} and a Binary Cross-Entropy loss of \textbf{0.7938}. We looked into a large number of attention maps over the images but we couldn't find any evidence that the model finds correlation between the similar regions of the images in the advertisements. We believe that this is due to the small number of epochs we used to train our model. 
\section{Comparison of the fusion modules}
\begin{figure}[h]
    \centering
    \includegraphics[width=\textwidth]{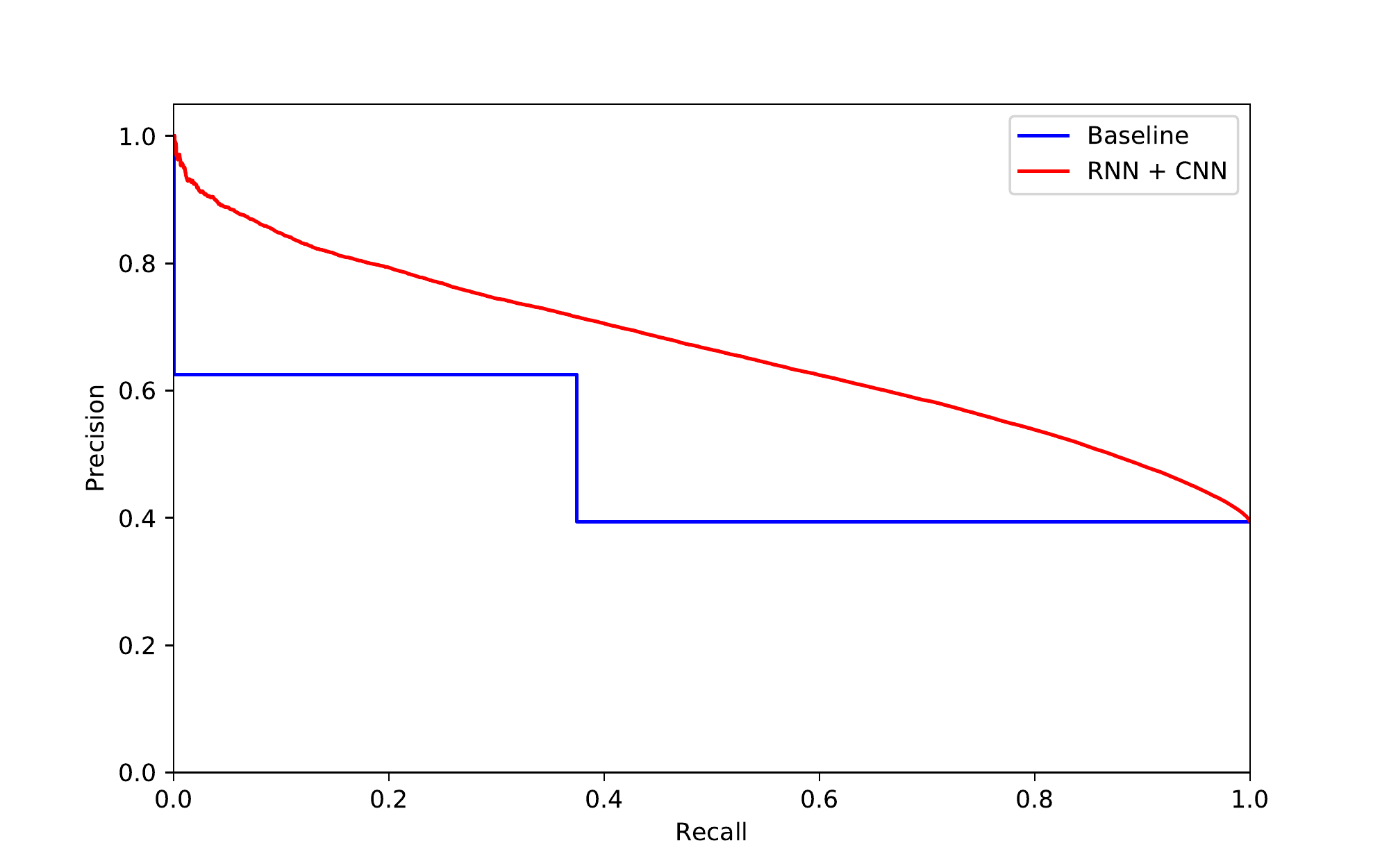}
    \caption{Comparison between the Precision - Recall curves of the Baseline model and the RNN + CNN model.}
    \label{fig:comparison}
\end{figure}

In this section, we compare the three models we have experimented with so far. We analyze them based on the average precision scores that they achieve on the test set, the set that we didn't use to choose the hyperparameter setting for the models that needed hyperparameter tuning. In Table \ref{ref:finaltable} we can see the average precision scores for each proposed solution. The RNN + CNN fusion module along with the Siamese Neural Network significantly outperform the baseline model. The Stacked Attention Network performs worse than the RNN + CNN module and we believe that this is because the model wasn't trained long enough. Therefore in the further discussion, we will omit the Stacked Attention Network module since it doesn't provide additional information to what we already have with the RNN + CNN model. As a way to illustrate the difference in performance between the baseline and the RNN + CNN model, we have plotted the Precision-Recall curve for both models in Figure \ref{fig:comparison} and as expected the PR curve of the RNN + CNN model has a much larger area under the curve then the PR curve of the baseline model.
\begin{table}[h!]
\centering
 \begin{tabular}{|c|c|} 
 \hline
 Model name & Average Precision Score on Test set \\ [0.5ex] 
 \hline
 Baseline model& 0.48 \\ [0.5ex] 
 \hline
 RNN + CNN model& $0.683 \pm 0.004$ \\ [0.5ex] 
 \hline
 SAN model& 0.574 \\ [0.5ex] 
 \hline
 \end{tabular}
 \caption{In this table we have presented the Average Precision Score that each model has achieved on the test set.}
 \label{ref:finaltable}
\end{table}

\begin{figure}[h]
    \centering
    \includegraphics[width=\textwidth]{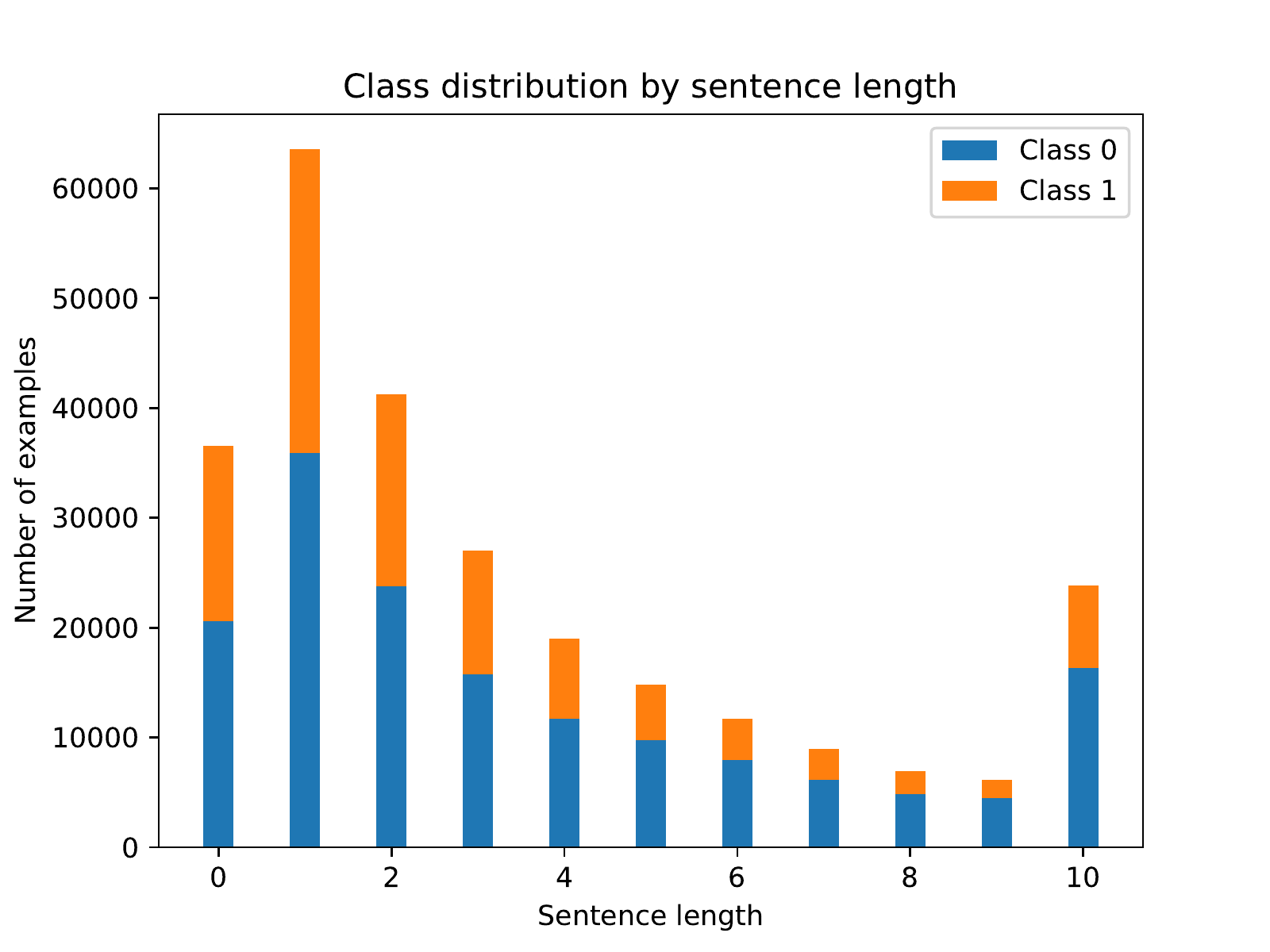}
    \caption{This bar chart shows the class distribution of each bucket.}
    \label{fig:buck_dist}
\end{figure}

Advertisements tend to have different description lengths, so we wanted to investigate how does the model perform when both descriptions in the pair are short, when the descriptions are both very long and when the descriptions are different lengths. We want to see if the models can predict equally well when the advertisements have different lengths because that would mean that the fusion module can capture crucial similarities between the descriptions even though, one description might contain more information then the other. To do that we computed the average length of each pair of descriptions. The maximum average could be 100 (because we limited our descriptions to 100 words) and the minimum could be 1, so we decided to divide the test set into 11 groups. The first group would be pairs that have an average description length of 1 to 9 words, the second group would have a length of 10 to 19 words, and so on until the last group, which is consisted of pairs that have an average length of 100 words. This is because we cut many descriptions in our preprocessing so, this group represents all the long advertisements. Table \ref{ref:finaltable2} shows the number of pairs that each group contains and Figure \ref{fig:buck_dist} shows the class distribution within each bucket.
\begin{table}[h!]
\centering
 \begin{tabular}{|c|c|c|c|c|c|c|c|c|c|c|} 
 \hline
 0 & 1 & 2 & 3 & 4 & 5 & 6 & 7 & 8 & 9 & 10 \\
 \hline
 36531&63573&41254&27039&18984&14836&11684&8925&6910&6144&23809\\
 \hline
 \end{tabular}
 \caption{Distribution of the test set pairs in the buckets of average description length. In the first row show the number of the bucket and in the second row we show how many pairs does the bucket contain.}
 \label{ref:finaltable2}
\end{table}

\begin{figure}[h]
    \centering
    \includegraphics[width=\textwidth]{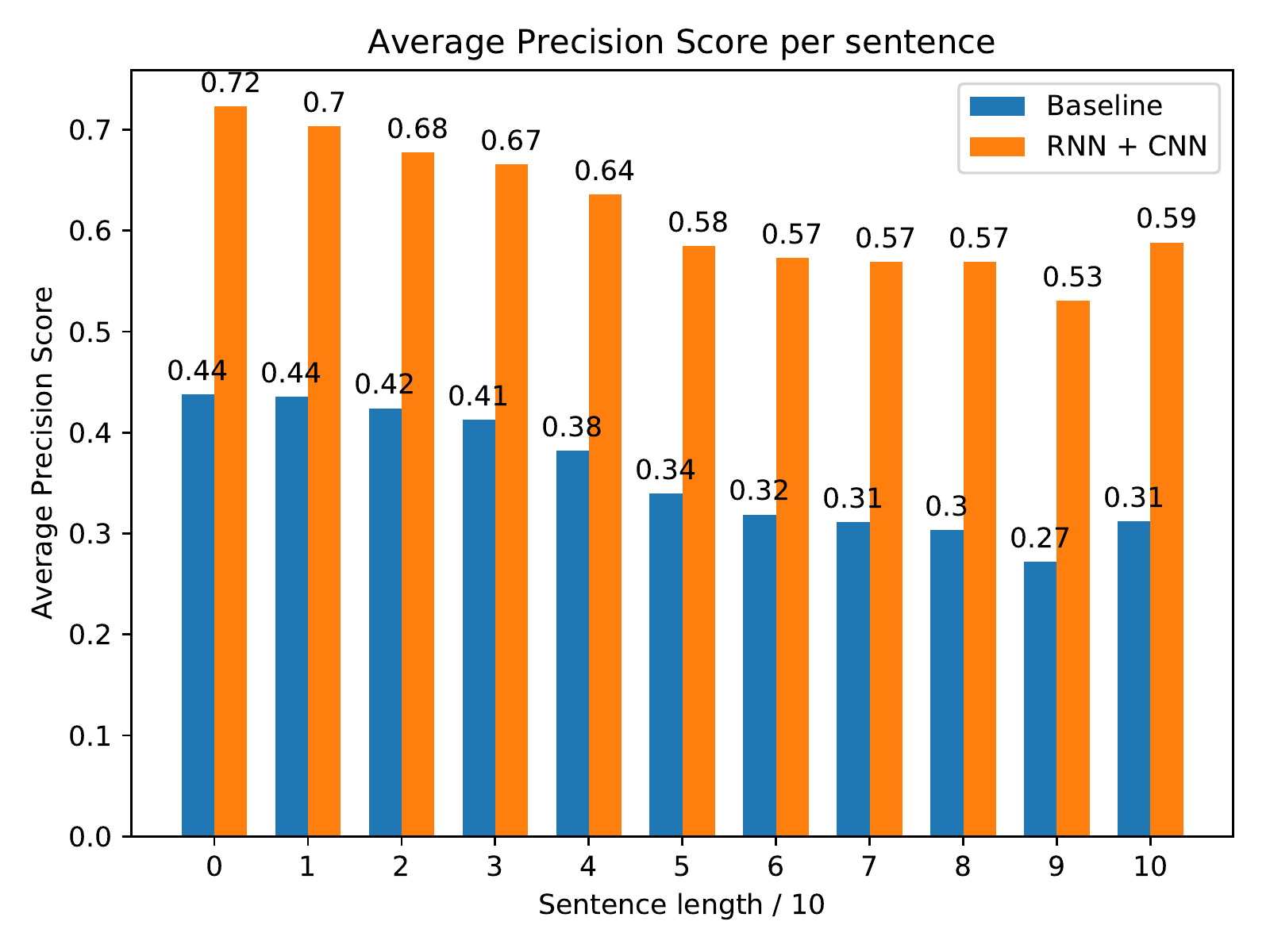}
    \caption{The bar chart show how do the models perform when they are predicting examples from the given buckets.}
    \label{fig:sent_figure}
\end{figure}
From Figure \ref{fig:sent_figure} we can see that both models perform better on pairs whose average description length is shorter than 40 words. The decline in the average precision score for pairs who have a bigger average description length maybe because there are pairs where the descriptions vary a lot in terms of word count. This means that both models predict poorly (worse than the models overall average) when, for example, one advertisement has a description of length 20 and the other advertisement has a description of length 100. There also seems to be a correlation between the distribution ratio and the average precision score. When the ration is closer to 1:3 instead of 1:2 in favour of the negative class, the model seems to have a lower average precision score. From Figure \ref{fig:sent_figure} we can see that most of the examples in the test set have an average length between 1 and 40, and the ratio of that joint bucket is much closer to 1:2 than it is to 1:3. This means that the model has learned to maximise the distribution of pairs with an average description length that is between 1 and 40. This results in a model that is too optimistic when it comes to advertisements that have very different or very big description lengths, therefore the model is more prone to classify a pair of dissimilar advertisements as similar. With this the number of \textit{FN} (false negatives) would increase, resulting in a smaller average precision score across the pairs that have a average description length bigger than 40 words. From Figure \ref{fig:sent_figure} we can see that the RNN + CNN fusion module outperforms the baseline model for every average description length, but it doesn't appear that it solves the problem of variable description lengths in the pair of advertisements.

It is hard for us to evaluate what kind of role does the image have in the improvement of the results. If we trained a separate model that only used the descriptions to match the advertisements we would have a clearer view of how does the inclusion of images in the fusion model affect the overall performance of the RNN + CNN solution. 

Overall, we proved that the neural network model performs better than the baseline model, but we couldn't distinguish which part of the fusion module contributes more to the improvement. We failed to prove that by increasing the complexity of the fusion module (upgrading from the RNN + CNN module to the Stacke Attention Network module) we would achieve better results and we also failed to train the Stacked Attention Network solution to find similar regions in the images of similar advertisements. We believe that with further training of the Stacked Attention Network solution, we might get better results and that by using the attention maps of the images, we would be able to have a better representation of the effect of the visual information. 
\chapter{Conclusions}
In this dissertation, we proposed two novel neural network-based solutions for the problem of multi-modal Record Linkage. When doing experiments with the RNN + CNN solution we found that the hyperparameters that were the best setting for the fusion module in \cite{antol2015vqa} and the Siamese Neural Network in \cite{koch2015siamese}, didn't achieve good results, therefore we concluded that the hyperparameter settings from these papers didn't transfer well from their tasks to ours. We were not able to achieve the objective set regarding the Stacked Attention Network solution, because of time constraints with our training process. We weren't able to find any evidence in the attention maps produced by the SAN solution that the model learned how to attend similar regions in the images of similar advertisements. By comparing the models we proved that the RNN + CNN solution achieves better Average Precision Score than the baseline. We also found that there is a correlation between how well the model performs depending on the average length of the descriptions of the advertisements. If the average length is over 40 words, then the model performs worse than the overall average precision score. The RNN + CNN is too optimistic with pairs of advertisements whose average description length is over 40 words, meaning that it usually classifies dissimilar pairs as similar. Therefore we believe that the RNN + CNN model struggles when the advertisements have very different description lengths or very long description lengths. 

For future work, we propose training the Stacked Attention Network for longer and training just an RNN module that will work only with the descriptions of the advertisements. We believe that by doing this, we could further investigate the overall effect of the images in our multi-modal fusion solutions. Another possibility would be to explore bilinear fusion as an alternative to the Stacked Neural Network fusion module \cite{cadene2019murel}.
% \section{Final Reminder}

% The body of your dissertation, before the references and any appendices,
% \emph{must} finish by page~40. The introduction, after preliminary material,
% should have started on page~1.

% You may not change the dissertation format (e.g., reduce the font
% size, change the margins, or reduce the line spacing from the default
% 1.5 spacing). Over length or incorrectly-formatted dissertations will
% not be accepted and you would have to modify your dissertation and
% resubmit.  You cannot assume we will check your submission before the
% final deadline and if it requires resubmission after the deadline to
% conform to the page and style requirements you will be subject to the
% usual late penalties based on your final submission time.

\bibliographystyle{plain}
\bibliography{main}

%% You can include appendices like this:
% \appendix
% 
% \chapter{First appendix}
% 
% \section{First section}
% 
% Markers do not have to consider appendices. Make sure that your contributions
% are made clear in the main body of the dissertation (within the page limit).

\end{document}